\documentclass[10pt,journal,compsoc]{IEEEtran}
\usepackage{cite}
\usepackage{graphicx}
\usepackage{amsmath}
\usepackage{algorithmic}
\usepackage{array}
\usepackage{fixltx2e}
\let\MYoriglatexcaption\caption
\renewcommand{\caption}[2][\relax]{\MYoriglatexcaption[#2]{#2}}
\usepackage{url}

\usepackage{xcolor}

\usepackage{epsfig}
\usepackage{epsfig}
\usepackage{graphicx}
\usepackage{amsmath}
\usepackage{amssymb}
\usepackage{bm}
\usepackage{color}
\usepackage{float}
\usepackage{stfloats}
\usepackage{caption}
\usepackage{subcaption}
\usepackage{algorithmic,algorithm}
\usepackage{mathrsfs}
\usepackage{multirow}
\usepackage[misc]{ifsym}

\usepackage[english]{babel} 
\usepackage{algorithm}
\usepackage{algorithmic}
\usepackage{xspace}
\usepackage{ragged2e}

\usepackage{gensymb} 
\usepackage{amsmath}
\usepackage{amsthm}
\usepackage{amsfonts}
\usepackage{mathrsfs}
\usepackage{amssymb}
\usepackage{tikz}
\usepackage{tikz-3dplot}
\usepackage{xspace}
\usepackage{makecell}
\newcommand*{\MinNumber}{0}
\newcommand*{\MidNumber}{12} 
\newcommand*{\MaxNumber}{40}
\newcommand*{\Ratio}{70}

\usepackage{pgfplots}
\usepgfplotslibrary{dateplot}
\newcommand{\cl}[1]{%
        \ifdim #1 pt > \MidNumber pt
            \pgfmathsetmacro{\PercentColor}{max(min(\Ratio*(#1 - \MidNumber)/(\MaxNumber-\MidNumber),\Ratio),0.00)} %
            \edef\x{\noexpand\cellcolor{red!\PercentColor!yellow!80}}\x #1
        \else
            \pgfmathsetmacro{\PercentColor}{max(min(\Ratio*(\MidNumber - #1)/(\MidNumber-\MinNumber),\Ratio),0.00)} %
            \edef\x{\noexpand\cellcolor{green!\PercentColor!yellow!80}}\x #1
        \fi
}

\usepackage{setspace}
\usepackage{colortbl}
\usepackage{url}
\usepackage{bbding}

\usepackage{cite}
\usepackage{booktabs}
\usepackage{footnote}
\usepackage[labelformat=simple]{subcaption}
\definecolor{maroon}{rgb}{.5,0,0}

\usepackage{hyperref}
\hypersetup{
    colorlinks=true,
    linkcolor=red,
    filecolor=magenta,      
    urlcolor=purple,
    citecolor=green,
}

\makeatletter
\DeclareRobustCommand\onedot{\futurelet\@let@token\@onedot}
\def\@onedot{\ifx\@let@token.\else.\null\fi\xspace}
\def\eg{\emph{e.g}\onedot} 
\def\ie{\emph{i.e}\onedot} 
\def\cf{\emph{c.f}\onedot} 
\def\etc{\emph{etc}\onedot} 
 
\def\etal{\emph{et al}\onedot}
\makeatother

\newcommand\customparagraph[1]{\vspace{0.4em}\noindent\textbf{#1}}

\newcommand{\Tref}[1]{Table~\ref{#1}}

\newcommand{\Fref}[1]{Figure~\ref{#1}}

\newcommand{\fref}[1]{Fig.~\ref{#1}}

\begin{document}

\title{EventAid: Benchmarking Event-aided Image/Video Enhancement Algorithms with Real-captured Hybrid Dataset}

\author{Peiqi Duan$^\dagger$, Boyu Li$^\dagger$, Yixin Yang, Hanyue Lou, Minggui Teng, Yi Ma, Boxin Shi$^\ddagger$,~\IEEEmembership{Senior Member,~IEEE}\\

\IEEEcompsocitemizethanks{

\IEEEcompsocthanksitem $^\dagger$ Contributed equally to this work as first authors.
\IEEEcompsocthanksitem $^\ddagger$ Corresponding author: shiboxin@pku.edu.cn
\IEEEcompsocthanksitem P. Duan, B. Li, Y. Yang, H. Lou, M. Teng, Y. Ma, and B. Shi are with National Key Laboratory for Multimedia Information Processing and National Engineering Research Center of Visual Technology, School of Computer Science, Peking University.
\IEEEcompsocthanksitem Project page: \textit{\url{ https://sites.google.com/view/EventAid-benchmark}}}
}

\markboth{IEEE TRANSACTIONS ON PATTERN ANALYSIS AND MACHINE INTELLIGENCE}
{Duan \MakeLowercase{\textit{et al.}}: Benchmarking Event-aided High-quality Image/Video Enhancement Algorithms with Real-captured Hybrid Dataset}

\IEEEtitleabstractindextext{%
\begin{abstract}\justifying
    Event cameras are emerging imaging technology that offers advantages over conventional frame-based imaging sensors in dynamic range and sensing speed. Complementing the rich texture and color perception of traditional image frames, the hybrid camera system of event and frame-based cameras enables high-performance imaging. With the assistance of event cameras, high-quality image/video enhancement methods make it possible to break the limits of traditional frame-based cameras, especially exposure time, resolution, dynamic range, and frame rate limits. This paper focuses on five event-aided image and video enhancement tasks (\ie{}, event-based video reconstruction, event-aided high frame rate video reconstruction, image deblurring, image super-resolution, and high dynamic range image reconstruction), provides an analysis of the effects of different event properties, a real-captured and ground truth labeled benchmark dataset, a unified benchmarking of state-of-the-art methods, and an evaluation for two mainstream event simulators. In detail, this paper collects a real-captured evaluation dataset \textsc{EventAid} for five event-aided image/video enhancement tasks, by using ``Event-RGB'' multi-camera hybrid system, taking into account scene diversity and spatiotemporal synchronization. We further perform quantitative and visual comparisons for state-of-the-art algorithms, provide a controlled experiment to analyze the performance limit of event-aided image deblurring methods, and discuss open problems to inspire future research.
\end{abstract}

}

\maketitle
\IEEEdisplaynontitleabstractindextext

\IEEEpeerreviewmaketitle

\IEEEraisesectionheading{\section{Introduction}\label{sec:introduction}}

\IEEEPARstart {E}{vent} cameras, also known as Dynamic Vision Sensors (DVS) \cite{dvs128, davis346}, draw on the perception mechanism of the human retina \cite{osti_Nature} to sense brightness changes in the scene in the form of ``event'' signals \cite{survey, davis240, davis346, dvs128}. Each pixel of the event camera compares the current and last light intensity state on a logarithmic scale and triggers a binary form event when the intensity variation exceeds the preset threshold \cite{dvs128, celex-v, survey}. Such a trigger mechanism enables event cameras to high-speed ($\sim 10\mu s$) perceive dynamic visual scenarios with a high dynamic range (HDR) ($\sim 120dB$) while lacking the absolute radiance intensity recording and static sensing \cite{davis240}. With the superior properties of HDR, low latency, and low redundancy \cite{dvs128}, event cameras make it possible to break through the bottlenecks of computer vision and robotic technologies based on traditional frame-based cameras. 

Thus far, event cameras have shown promising capability in solving classical as well as new computer vision and robotics tasks, including low-level tasks such as high frame-rate (HFR) video synthesis \cite{e2vid-pami19, ssl-e2vid} and HDR image reconstruction \cite{reinbacher2016realtime, Chane2016tonemap}, middle-level tasks such as optical flow \cite{zhu2018evflownet, snn-opt-flow-eccv20} and scene depth estimation \cite{zhu2019unsupervised, cont-max-cvpr18}, and high-level tasks such as 3D reconstruction \cite{rebecq2018emvs, baudron2020e3d}, object tracking \cite{stnet, visevent}, object detection \cite{ramesh2020boosted, ramesh2020low}, SLAM \cite{vidal2018ultimate}, and autonomous wheel steering \cite{Autonomous}.

\begin{figure*}[t]
    \centering
    \includegraphics[width=\linewidth]{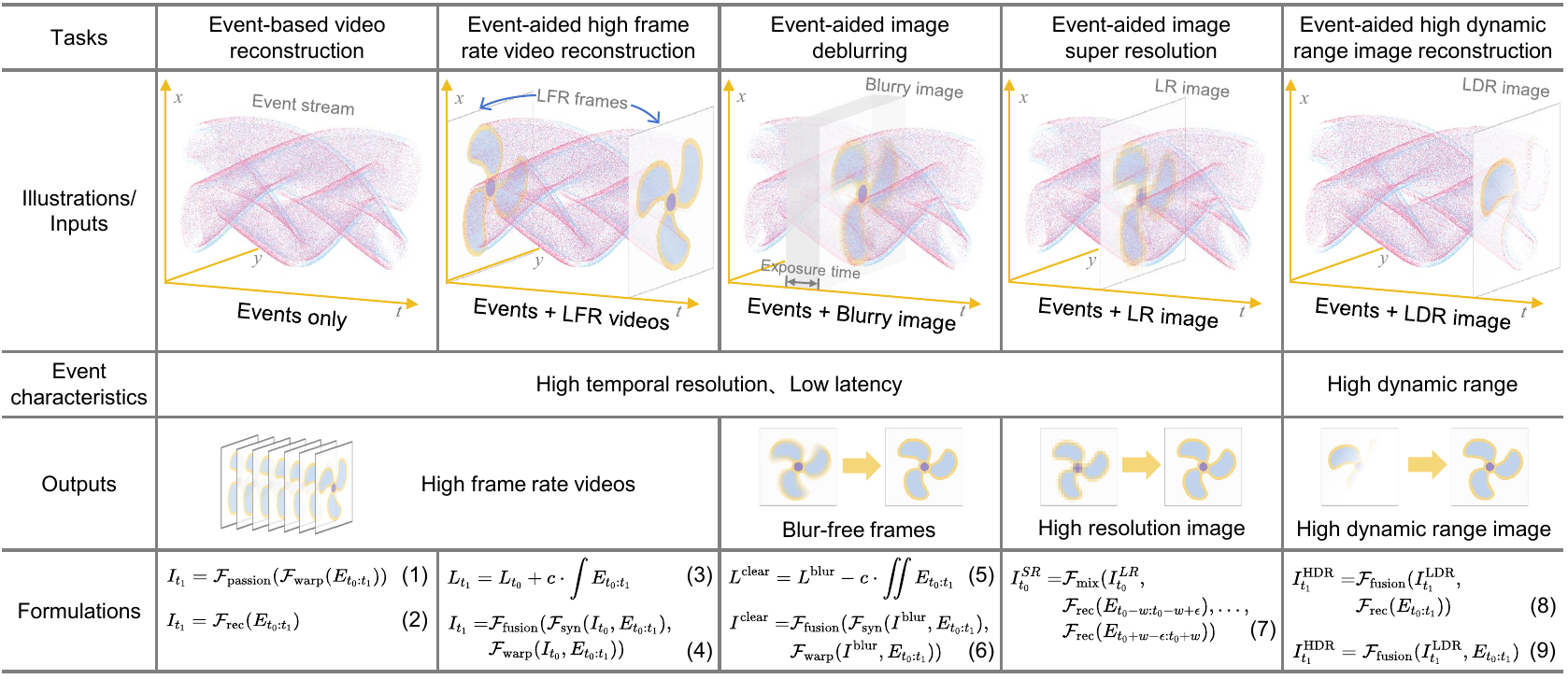}
    \caption{\textit{1st row}: Five event-aided image/video enhancement tasks. \textit{2nd row}: The illustrations of different tasks, when the event camera and the RGB camera shoot rotating fan blades at the same time and in the same field of view, the event camera asynchronously triggers positive (blue dot) and negative (red dot) events with high temporal resolution, while the RGB camera outputs images frame by frame. \textit{3rd row}: The Event characteristics that each task used to break through the performance bottlenecks limited by frames. \textit{4th row}: Outputs of different tasks. \textit{5th row}: Formulations of each task.}
    \label{fig:overview}
\end{figure*}

Due to the special triggering mechanism and much shorter research times compared to frame-based cameras, the signal quality of event cameras degrades when the scenes are relatively static, and suffers from severe noises and poor color perception \cite{NeuroZoom}. By contrast, traditional frame-based RGB cameras are the mainstream sensors of computer vision and robotic technologies that feature rich color, texture, and semantic information as well as lower noise. Witnessing and experiencing the success of frame-based RGB sensors and corresponding algorithms over the past decades, researchers have built large-scale datasets \cite{imagenet, coco}, various well-designed network architectures \cite{vaswani2017attention, 3dunet20163DUL}, and even foundation models \cite{Ko_2023_foundationModel, Yu_2023_foundationModel} for frame-based vision. Such sensory motivates researchers to leverage the complementary advantages of both ends through an ``Event-RGB'' hybrid multi-camera fusion \cite{gef-tpami, Timelens} and to use the existing achievements of frame-based cameras for accelerating the research of event cameras. This fusion has been extensively explored in the field of high-quality imaging. To take advantage of the high speed and HDR features of event cameras, break through the traditional imaging bottlenecks, and meet the image quality requirements of human and machine vision, researchers have bridged the event and image modality in recent years \cite{gef-tpami, e2vid-cvpr19, e2sri-cvpr20, edi-cvpr19, Timelens, timelens++, han2020hdr, EvUnroll_cvpr22}. We classify such tasks as \textbf{event-aided} (also known as event-guided \cite{gef-cvpr20, Kim2021Eventguided, Yang_2023_HDR}) \textbf{image/video enhancement} methods, and we focus on five event-aided tasks in this paper (as shown in \fref{fig:overview}).

To benchmark event-aided image/video enhancement methods with real data, a real ``Event-RGB'' hybrid multi-camera system with high-precision spatiotemporal synchronization is necessary. Currently, mainly three types of ``Event-RGB'' hybrid multi-camera systems are in use: (1) As early attempts, ATIS \cite{atis} and DAVIS \cite{davis240, davis346} cameras embed intensity-recording subpixels or conventional Active Pixel Sensors (APS) with event sensors to ``Event-RGB'' simultaneous imaging. This kind of design is the optimal way to achieve multi-camera spatiotemporal synchronization, while there are bottlenecks that lead to severe noise and low resolution \cite{NeuroZoom} (\eg{}, the resolution of the widely-used DAVIS346 \cite{davis346} is only $346\times260$). (2) To match high-quality frame cameras, building an event camera and a frame camera into a dual-camera system becomes another option \cite{Timelens}. Nevertheless, the different homography matrices of different scene depths make it difficult for this dual-camera system to avoid spatial matching errors. (3) Wang \etal \cite{gef-cvpr20} and Han \etal \cite{han2020hdr} first propose a hybrid camera system that physically co-located an event camera and a frame camera via a beam splitter, with two cameras sharing a common field of view. Although there are still problems of non-portability and unstable matching accuracy, such a system can easily replace cameras while avoiding field-of-view misalignment caused by binocular disparity. Based on the above three hybrid systems, many evaluation datasets for event-aided image/video enhancement methods have been proposed (as listed in \Tref{tab:summary}). To avoid the imperfect conditions of the above real systems, using event simulators to generate events and prepare datasets is also a popular choice \cite{ev_deblur, Timelens}, while the unavoidable gap between real-captured and simulated data \cite{NeuroZoom} (real-sim gap) makes the benchmark results less convincing to reveal the real performance of algorithms. As event-aided image/video enhancement methods are continuously springing up, a high-quality real-captured dataset for comprehensive benchmarking them on a unified scale is urgently needed.

In this paper, we propose a real-world and comprehensive evaluation dataset, named \textsc{EventAid}, for evaluating five mainstream event-aided image/video enhancement tasks, taking into account spatial alignment and temporal synchronization of two sensors as well as scene diversity. All data, including input events and frames and the ground truth, are real-captured by beam splitter-mounted hybrid camera systems. The sub-datasets corresponding to each task are: \textsc{EventAid-R} for event-based video reconstruction, \textsc{EventAid-F} for HFR video reconstruction, \textsc{EventAid-B} for image deblurring, \textsc{EventAid-S} for image super-resolution (SR), and \textsc{EventAid-D} for HDR image reconstruction. We benchmark a total of 19 state-of-the-art algorithms for all five tasks and statistically analyze their performances in different scenarios. To evaluate the real-sim gap of event simulators, we generate simulated datasets with different simulators and execute the benchmarking again. To the best of our knowledge, this is the first high-quality benchmark dataset for event-aided image/video enhancement tasks with real-captured data that allow quantitative and qualitative evaluations, and the first comprehensive benchmarking of existing methods in diverse real-world scenes with unified evaluation protocols.

This paper makes the following contributions:
\begin{itemize}
    \item We categorize five mainstream event-aided image/video enhancement tasks using unified formulations, as well as summarize and compare the existing evaluation datasets.
    \item We collect the first real-captured dataset \textsc{EventAid} to evaluate the performance of existing methods of five tasks, with high accuracy of spatiotemporal synchronization between two sensors as well as great scene diversity (35 scenes in total).
    \item We benchmark a total of 19 state-of-the-art methods for all five tasks and statistically analyze the performance of them. We further compare and evaluate the real-sim gap of two widely used event simulators by referring to the real data benchmark results.
    \item Based on our benchmark evaluation, open problems from different aspects of these five tasks, such as evaluation metric, feature extraction, artifacts suppression, color restoration, and so on, are discussed to inspire future research.
\end{itemize}


\section{Event-aided imaging categorization}
This section first presents the mathematical form of the event trigger model and categorizes and formulates existing event-aided image/video enhancement methods into five tasks with unified notations. Then, we list, organize, and summarize existing evaluation datasets for these five tasks.

\subsection{Event-aided imaging model formulation}

We first formulate the event trigger model and build the relationship to the image-based counterpart. Consider a latent spatiotemporal volume in which an intensity field is sampled by an ideal frame-based camera that can output blur-free, high-resolution, and HDR intensity images $I_t$ at any moment. The event output at ${t_0}$ can be described as:

\begin{equation}
    E_{t_0} = \Gamma\big\{\log(\frac{I_{t_0}+b}{I_{t_0-1}+b}), c+n_\text{event}\big\}, \tag{10}
    \label{eq:event-noiseless}
\end{equation}
where $\Gamma\{\theta, c\}$ represents the conversion function from log-intensity to events, and $b$ is an offset value to prevent $\log(0)$. $\Gamma\{\theta, c\} = 1$ when $\theta\geq c$, indicating a positive event; $\Gamma\{\theta, c\} = -1$ when $\theta\leq-c$, indicating a negative event; and $\Gamma\{\theta, c\} = 0$ when $|\theta|<c$, indicating that no event has been fired. $n_\text{event}$ represents the perturbation noise pivoted at the firing threshold $c$. The duration from ${t_0-1}$ to ${t_0}$ corresponds to the minimum response delay of the event camera. The dead pixels can be interpreted as $c$ being significantly low or high. In event-aided image/video enhancement tasks, $I_t$ is taken as the ground truth.

Here we model five tasks and compare their relationships and differences. \Fref{fig:overview} shows the illustrations of input and process for each task as well as the equations to be solved for each task.

\begin{table*}[htbp]
\renewcommand\arraystretch{1.4}
\LARGE
  \centering
  \caption{The summary of existing evaluation datasets of five event-aided image/video enhancement tasks. The following four characteristics as we marked in the 1st row are compared: 1. Real-captured data: the input images and events, and ground truth are real-captured or simulated. 2. Spatiotemporal synchronization of two sensors. 3. Event/Frame-based sensor: the spatial resolution, color imaging type, and frame rate parameters of cameras. 4. Scene diversity. (``-'' represents a ``not applied'' attribute, \eg{}, the event-based video reconstruction does not require input images. The resolution and frame rate of some datasets are not completely consistent, we use ``$\sim$'' and ``$\textless$'' to represent the approximate value distribution.}
  
    \resizebox{\hsize}{!}{
    \scalebox{1}{
    \begin{tabular}{c|c|l|c|c|c|c|c|c|c|c|c|r|c|c|c|c}
    \toprule[3pt]
    \multirow{2}[4]{*}{} & \multicolumn{1}{c|}{\multirow{2}[4]{*}[-0.7em]{\textbf{\makecell[c]{Simulated\\or\\real dataset}}}} & \multicolumn{1}{c|}{\multirow{2}[4]{*}[-0.7em]{\textbf{Dataset name}}} & \multicolumn{3}{c|}{\textbf{Real-captured data}} & \multicolumn{2}{c|}{\textbf{Spatiotemporal synchronization of two sensors}} & \multicolumn{2}{c|}{\textbf{Event sensor}} & \multicolumn{3}{c|}{\textbf{Frame-based sensor}} & \multicolumn{4}{c}{\textbf{Scene diversity}} \\
\cmidrule{4-17}          &       &       & \multicolumn{1}{c|}{\textbf{\makecell[c]{Input\\image}}} & \multicolumn{1}{c|}{\textbf{\makecell[c]{Input\\events}}} & \multicolumn{1}{c|}{\textbf{\makecell[c]{Ground\\truth}}} & \textbf{Spatial matching } & \multicolumn{1}{c|}{\textbf{\makecell[c]{Temporal\\synchronization}}} & \multicolumn{1}{c|}{\textbf{\makecell[c]{Camera model}}} & \multicolumn{1}{c|}{\textbf{\makecell[c]{Spatial\\resolution}}} & \multicolumn{1}{c|}{\textbf{\makecell[c]{Spatial\\resolution}}} & \multicolumn{1}{c|}{\textbf{\makecell[c]{Color /\\gray}}} & \multicolumn{1}{c|}{\textbf{\makecell[c]{Frame\\rate}}} & \multicolumn{1}{c|}{\textbf{\makecell[c]{Indoor+\\Outdoor}}} & \multicolumn{1}{c|}{\textbf{\makecell[c]{Ego+\\Local\\motion}}} & \multicolumn{1}{c|}{\textbf{\makecell[c]{Slow+\\Fast\\motion}}} & \multicolumn{1}{c}{\textbf{\makecell[c]{High\\texture}}} \\
    \midrule[2pt]
    \multicolumn{1}{c|}{\multirow{7}[14]{*}{\textbf{\makecell[c]{Event-based\\video reconstruction}}}} & \multicolumn{1}{c|}{\textbf{Simulation}} & \cellcolor[rgb]{ .839,  .863,  .894}EventNFS\cite{EventZoom} & \cellcolor[rgb]{ .839,  .863,  .894}\textbf{-} & \cellcolor[rgb]{ .839,  .863,  .894}\textcolor[rgb]{ 0,  .69,  .314}{\checkmark} & \cellcolor[rgb]{ .839,  .863,  .894}\textcolor[rgb]{ 1,  0,  0}{\XSolidBrush} & \cellcolor[rgb]{ .839,  .863,  .894}Display+camera calibration & \cellcolor[rgb]{ .839,  .863,  .894}Mark points matching & \cellcolor[rgb]{ .839,  .863,  .894}DAVIS346 mono & \cellcolor[rgb]{ .839,  .863,  .894}222$\times$124 & \cellcolor[rgb]{ .839,  .863,  .894}222$\times$124 & \cellcolor[rgb]{ .839,  .863,  .894}color & \cellcolor[rgb]{ .839,  .863,  .894}- & \cellcolor[rgb]{ .839,  .863,  .894}\textcolor[rgb]{ 0,  .69,  .314}{\checkmark} & \cellcolor[rgb]{ .839,  .863,  .894}\textcolor[rgb]{ 0,  .69,  .314}{\checkmark} & \cellcolor[rgb]{ .839,  .863,  .894}\textcolor[rgb]{ 0,  .69,  .314}{\checkmark} & \cellcolor[rgb]{ .839,  .863,  .894}\textcolor[rgb]{ 0,  .69,  .314}{\checkmark} \\
\cmidrule{2-17}          & \multicolumn{1}{c|}{\multirow{6}[12]{*}{\textbf{Real}}} &   IJRR\cite{davis} & \textbf{-} & \textcolor[rgb]{ 0,  .69,  .314}{\checkmark} & \textcolor[rgb]{ 0,  .69,  .314}{\checkmark} & Frame+event sensor & Chip synchronization & DAVIS240 & 240$\times$180 & 240$\times$180 & gray  & $\sim$24 FPS & \textcolor[rgb]{ 0,  .69,  .314}{\checkmark} & \textcolor[rgb]{ 0,  .69,  .314}{\checkmark} & \textcolor[rgb]{ 0,  .69,  .314}{\checkmark} & \textcolor[rgb]{ 0,  .69,  .314}{\checkmark} \\
\cmidrule{3-17}          &       &   HQF\cite{sim-to-real-gap-eccv20} & \textbf{-} & \textcolor[rgb]{ 0,  .69,  .314}{\checkmark} & \textcolor[rgb]{ 0,  .69,  .314}{\checkmark} & Frame+event sensor & Chip synchronization & DAVIS240 & 240$\times$180 & 240$\times$180 & gray  & $\textless$ 30 FPS & \textcolor[rgb]{ 0,  .69,  .314}{\checkmark} & \textcolor[rgb]{ 0,  .69,  .314}{\checkmark} & \textcolor[rgb]{ 0,  .69,  .314}{\checkmark} & \textcolor[rgb]{ 0,  .69,  .314}{\checkmark} \\
\cmidrule{3-17}          &       &   DVS-Dark\cite{ev-sid-eccv20} & \textbf{-} & \textcolor[rgb]{ 0,  .69,  .314}{\checkmark} & \textcolor[rgb]{ 0,  .69,  .314}{\checkmark} & Frame+event sensor & Chip synchronization & DAVIS240 & 240$\times$180 & 240$\times$180 & gray  & $\textless$ 30 FPS & \textcolor[rgb]{ 0,  .69,  .314}{\checkmark} & \textcolor[rgb]{ 1,  0,  0}{\XSolidBrush} & \textcolor[rgb]{ 1,  0,  0}{\XSolidBrush} & \textcolor[rgb]{ 1,  0,  0}{\XSolidBrush} \\
\cmidrule{3-17}          &       &   MVSEC\cite{mvsec-ral18} & \textbf{-} & \textcolor[rgb]{ 0,  .69,  .314}{\checkmark} & \textcolor[rgb]{ 0,  .69,  .314}{\checkmark} & Frame+event sensor & Chip synchronization & DAVIS346 mono & 346$\times$260 & 346$\times$260 & gray  & 50 FPS & \textcolor[rgb]{ 0,  .69,  .314}{\checkmark} & \textcolor[rgb]{ 0,  .69,  .314}{\checkmark} & \textcolor[rgb]{ 0,  .69,  .314}{\checkmark} & \textcolor[rgb]{ 1,  0,  0}{\XSolidBrush} \\
\cmidrule{3-17}          &       &   CED\cite{ced} & \textbf{-} & \textcolor[rgb]{ 0,  .69,  .314}{\checkmark} & \textcolor[rgb]{ 0,  .69,  .314}{\checkmark} & Frame+event sensor & Chip synchronization & DAVIS346 color & 346$\times$260 & 346$\times$260 & color & $\textless$ 50 FPS & \textcolor[rgb]{ 0,  .69,  .314}{\checkmark} & \textcolor[rgb]{ 0,  .69,  .314}{\checkmark} & \textcolor[rgb]{ 1,  0,  0}{\XSolidBrush} & \textcolor[rgb]{ 1,  0,  0}{\XSolidBrush} \\
\cmidrule{3-17}          &       & \cellcolor[rgb]{ 1,  .953,  .792}\textbf{\textsc{EventAid-R}} & \cellcolor[rgb]{ 1,  .953,  .792}\textbf{-} & \cellcolor[rgb]{ 1,  .953,  .792}\textcolor[rgb]{ 0,  .69,  .314}{\checkmark} & \cellcolor[rgb]{ 1,  .953,  .792}\textcolor[rgb]{ 0,  .69,  .314}{\checkmark} & \cellcolor[rgb]{ 1,  .953,  .792}Beam splitter & \cellcolor[rgb]{ 1,  .953,  .792}External clock triggering & \cellcolor[rgb]{ 1,  .953,  .792}Prophesee & \cellcolor[rgb]{ 1,  .953,  .792}$\sim$954$\times$636 & \cellcolor[rgb]{ 1,  .953,  .792}$\sim$954$\times$636 & \cellcolor[rgb]{ 1,  .953,  .792}color & \cellcolor[rgb]{ 1,  .953,  .792}150 FPS & \cellcolor[rgb]{ 1,  .953,  .792}\textcolor[rgb]{ 0,  .69,  .314}{\checkmark} & \cellcolor[rgb]{ 1,  .953,  .792}\textcolor[rgb]{ 0,  .69,  .314}{\checkmark} & \cellcolor[rgb]{ 1,  .953,  .792}\textcolor[rgb]{ 0,  .69,  .314}{\checkmark} & \cellcolor[rgb]{ 1,  .953,  .792}\textcolor[rgb]{ 0,  .69,  .314}{\checkmark} \\
    \midrule[2pt]
    \multicolumn{1}{c|}{\multirow{7}[14]{*}{\textbf{\makecell[c]{Event-aided\\high frame rate\\video reconstruction}}}} & \multicolumn{1}{c|}{\multirow{2}[4]{*}{\textbf{Simulation}}} & \cellcolor[rgb]{ .839,  .863,  .894}Tulyakov \etal\cite{Timelens} & \cellcolor[rgb]{ .839,  .863,  .894}\textcolor[rgb]{ 0,  .69,  .314}{\checkmark} & \cellcolor[rgb]{ .839,  .863,  .894}\textcolor[rgb]{ 1,  0,  0}{\XSolidBrush} & \cellcolor[rgb]{ .839,  .863,  .894}\textcolor[rgb]{ 0,  .69,  .314}{\checkmark} & \cellcolor[rgb]{ .839,  .863,  .894}- & \cellcolor[rgb]{ .839,  .863,  .894}- & \cellcolor[rgb]{ .839,  .863,  .894}Simulation & \cellcolor[rgb]{ .839,  .863,  .894}1280$\times$720 & \cellcolor[rgb]{ .839,  .863,  .894}1280$\times$720 & \cellcolor[rgb]{ .839,  .863,  .894}color & \cellcolor[rgb]{ .839,  .863,  .894}- & \cellcolor[rgb]{ .839,  .863,  .894}\textcolor[rgb]{ 0,  .69,  .314}{\checkmark} & \cellcolor[rgb]{ .839,  .863,  .894}\textcolor[rgb]{ 0,  .69,  .314}{\checkmark} & \cellcolor[rgb]{ .839,  .863,  .894}\textcolor[rgb]{ 0,  .69,  .314}{\checkmark} & \cellcolor[rgb]{ .839,  .863,  .894}\textcolor[rgb]{ 0,  .69,  .314}{\checkmark} \\
\cmidrule{3-17}          &       & \cellcolor[rgb]{ .839,  .863,  .894}GoPro+ESIM\cite{esl-eccv20} & \cellcolor[rgb]{ .839,  .863,  .894}\textcolor[rgb]{ 0,  .69,  .314}{\checkmark} & \cellcolor[rgb]{ .839,  .863,  .894}\textcolor[rgb]{ 1,  0,  0}{\XSolidBrush} & \cellcolor[rgb]{ .839,  .863,  .894}\textcolor[rgb]{ 0,  .69,  .314}{\checkmark} & \cellcolor[rgb]{ .839,  .863,  .894}- & \cellcolor[rgb]{ .839,  .863,  .894}- & \cellcolor[rgb]{ .839,  .863,  .894}Simulation & \cellcolor[rgb]{ .839,  .863,  .894}1280$\times$720 & \cellcolor[rgb]{ .839,  .863,  .894}1280$\times$720 & \cellcolor[rgb]{ .839,  .863,  .894}color & \cellcolor[rgb]{ .839,  .863,  .894}240 FPS & \cellcolor[rgb]{ .839,  .863,  .894}\textcolor[rgb]{ 0,  .69,  .314}{\checkmark} & \cellcolor[rgb]{ .839,  .863,  .894}\textcolor[rgb]{ 0,  .69,  .314}{\checkmark} & \cellcolor[rgb]{ .839,  .863,  .894}\textcolor[rgb]{ 0,  .69,  .314}{\checkmark} & \cellcolor[rgb]{ .839,  .863,  .894}\textcolor[rgb]{ 0,  .69,  .314}{\checkmark} \\
\cmidrule{2-17}          & \multicolumn{1}{c|}{\multirow{5}[10]{*}{\textbf{Real}}} &   SloMo-DVS\cite{Yu_2021_ICCV} & \textcolor[rgb]{ 0,  .69,  .314}{\checkmark} & \textcolor[rgb]{ 0,  .69,  .314}{\checkmark} & \textcolor[rgb]{ 0,  .69,  .314}{\checkmark} & Frame+event sensor & Chip synchronization & DAVIS240 & 240$\times$180 & 240$\times$180 & gray  & $\textless$ 30 FPS & \textcolor[rgb]{ 0,  .69,  .314}{\checkmark} & \textcolor[rgb]{ 0,  .69,  .314}{\checkmark} & \textcolor[rgb]{ 0,  .69,  .314}{\checkmark} & \textcolor[rgb]{ 0,  .69,  .314}{\checkmark} \\
\cmidrule{3-17}          &       &   GEF\cite{gef-tpami} & \textcolor[rgb]{ 0,  .69,  .314}{\checkmark} & \textcolor[rgb]{ 0,  .69,  .314}{\checkmark} & \textcolor[rgb]{ 0,  .69,  .314}{\checkmark} & Beam splitter & Mark points matching & DAVIS240 & 190$\times$180 & 1520$\times$1440 & color & 20 FPS & \textcolor[rgb]{ 0,  .69,  .314}{\checkmark} & \textcolor[rgb]{ 0,  .69,  .314}{\checkmark} & \textcolor[rgb]{ 1,  0,  0}{\XSolidBrush} & \textcolor[rgb]{ 0,  .69,  .314}{\checkmark} \\
\cmidrule{3-17}          &       &   HS-ERGB\cite{Timelens} & \textcolor[rgb]{ 0,  .69,  .314}{\checkmark} & \textcolor[rgb]{ 0,  .69,  .314}{\checkmark} & \textcolor[rgb]{ 0,  .69,  .314}{\checkmark} &  Dual camera setup & External clock triggering & Prophesee & $\sim$900$\times$800 & $\sim$900$\times$800 & color & 226 FPS & \textcolor[rgb]{ 0,  .69,  .314}{\checkmark} & \textcolor[rgb]{ 0,  .69,  .314}{\checkmark} & \textcolor[rgb]{ 0,  .69,  .314}{\checkmark} & \textcolor[rgb]{ 0,  .69,  .314}{\checkmark} \\
\cmidrule{3-17}          &       &   BS-ERGB\cite{timelens++} & \textcolor[rgb]{ 0,  .69,  .314}{\checkmark} & \textcolor[rgb]{ 0,  .69,  .314}{\checkmark} & \textcolor[rgb]{ 0,  .69,  .314}{\checkmark} & Beam splitter & External clock triggering & Prophesee & 970$\times$625 & 970$\times$625 & color & 28 FPS & \textcolor[rgb]{ 0,  .69,  .314}{\checkmark} & \textcolor[rgb]{ 0,  .69,  .314}{\checkmark} & \textcolor[rgb]{ 0,  .69,  .314}{\checkmark} & \textcolor[rgb]{ 0,  .69,  .314}{\checkmark} \\
\cmidrule{3-17}          &       & \cellcolor[rgb]{ 1,  .953,  .792}\textbf{\textsc{EventAid-F}} & \cellcolor[rgb]{ 1,  .953,  .792}\textcolor[rgb]{ 0,  .69,  .314}{\checkmark} & \cellcolor[rgb]{ 1,  .953,  .792}\textcolor[rgb]{ 0,  .69,  .314}{\checkmark} & \cellcolor[rgb]{ 1,  .953,  .792}\textcolor[rgb]{ 0,  .69,  .314}{\checkmark} & \cellcolor[rgb]{ 1,  .953,  .792}Beam splitter & \cellcolor[rgb]{ 1,  .953,  .792}External clock triggering & \cellcolor[rgb]{ 1,  .953,  .792}Prophesee & \cellcolor[rgb]{ 1,  .953,  .792}$\sim$954$\times$636 & \cellcolor[rgb]{ 1,  .953,  .792}$\sim$954$\times$636 & \cellcolor[rgb]{ 1,  .953,  .792}color & \cellcolor[rgb]{ 1,  .953,  .792}150 FPS & \cellcolor[rgb]{ 1,  .953,  .792}\textcolor[rgb]{ 0,  .69,  .314}{\checkmark} & \cellcolor[rgb]{ 1,  .953,  .792}\textcolor[rgb]{ 0,  .69,  .314}{\checkmark} & \cellcolor[rgb]{ 1,  .953,  .792}\textcolor[rgb]{ 0,  .69,  .314}{\checkmark} & \cellcolor[rgb]{ 1,  .953,  .792}\textcolor[rgb]{ 0,  .69,  .314}{\checkmark} \\
    \midrule[2pt]
    \multicolumn{1}{c|}{\multirow{6}[12]{*}{\textbf{\makecell[c]{Event-aided\\image deblurring}}}} & \multicolumn{1}{c|}{\multirow{4}[8]{*}{\textbf{Simulation}}} & \cellcolor[rgb]{ .839,  .863,  .894}GoPro+ESIM\cite{ev_deblur} & \cellcolor[rgb]{ .839,  .863,  .894}\textcolor[rgb]{ 1,  0,  0}{\XSolidBrush} & \cellcolor[rgb]{ .839,  .863,  .894}\textcolor[rgb]{ 1,  0,  0}{\XSolidBrush} & \cellcolor[rgb]{ .839,  .863,  .894}\textcolor[rgb]{ 0,  .69,  .314}{\checkmark} & \cellcolor[rgb]{ .839,  .863,  .894}- & \cellcolor[rgb]{ .839,  .863,  .894}- & \cellcolor[rgb]{ .839,  .863,  .894}Simulation & \cellcolor[rgb]{ .839,  .863,  .894}1280$\times$720 & \cellcolor[rgb]{ .839,  .863,  .894}1280$\times$720 & \cellcolor[rgb]{ .839,  .863,  .894}color & \cellcolor[rgb]{ .839,  .863,  .894}$\sim$34 FPS & \cellcolor[rgb]{ .839,  .863,  .894}\textcolor[rgb]{ 0,  .69,  .314}{\checkmark} & \cellcolor[rgb]{ .839,  .863,  .894}\textcolor[rgb]{ 0,  .69,  .314}{\checkmark} & \cellcolor[rgb]{ .839,  .863,  .894}\textcolor[rgb]{ 0,  .69,  .314}{\checkmark} & \cellcolor[rgb]{ .839,  .863,  .894}\textcolor[rgb]{ 0,  .69,  .314}{\checkmark} \\
\cmidrule{3-17}          &       & \cellcolor[rgb]{ .839,  .863,  .894}Blur-DVS\cite{jiang2020learning} & \cellcolor[rgb]{ .839,  .863,  .894}\textcolor[rgb]{ 1,  0,  0}{\XSolidBrush} & \cellcolor[rgb]{ .839,  .863,  .894}\textcolor[rgb]{ 0,  .69,  .314}{\checkmark} & \cellcolor[rgb]{ .839,  .863,  .894}\textcolor[rgb]{ 0,  .69,  .314}{\checkmark} & \cellcolor[rgb]{ .839,  .863,  .894}Frame+event sensor & \cellcolor[rgb]{ .839,  .863,  .894}Chip synchronization & \cellcolor[rgb]{ .839,  .863,  .894}DAVIS240 & \cellcolor[rgb]{ .839,  .863,  .894}240$\times$180 & \cellcolor[rgb]{ .839,  .863,  .894}240$\times$180 & \cellcolor[rgb]{ .839,  .863,  .894}gray & \cellcolor[rgb]{ .839,  .863,  .894}$\sim$4 FPS & \cellcolor[rgb]{ .839,  .863,  .894}\textcolor[rgb]{ 1,  0,  0}{\XSolidBrush} & \cellcolor[rgb]{ .839,  .863,  .894}\textcolor[rgb]{ 0,  .69,  .314}{\checkmark} & \cellcolor[rgb]{ .839,  .863,  .894}\textcolor[rgb]{ 1,  0,  0}{\XSolidBrush} & \cellcolor[rgb]{ .839,  .863,  .894}\textcolor[rgb]{ 0,  .69,  .314}{\checkmark} \\
\cmidrule{3-17}          &       & \cellcolor[rgb]{ .839,  .863,  .894}RBE\cite{ev_deblur} & \cellcolor[rgb]{ .839,  .863,  .894}\textcolor[rgb]{ 1,  0,  0}{\XSolidBrush} & \cellcolor[rgb]{ .839,  .863,  .894}\textcolor[rgb]{ 0,  .69,  .314}{\checkmark} & \cellcolor[rgb]{ .839,  .863,  .894}\textcolor[rgb]{ 0,  .69,  .314}{\checkmark} & \cellcolor[rgb]{ .839,  .863,  .894}Frame+event sensor & \cellcolor[rgb]{ .839,  .863,  .894}Chip synchronization & \cellcolor[rgb]{ .839,  .863,  .894}DAVIS240 & \cellcolor[rgb]{ .839,  .863,  .894}240$\times$180 & \cellcolor[rgb]{ .839,  .863,  .894}240$\times$180 & \cellcolor[rgb]{ .839,  .863,  .894}gray & \cellcolor[rgb]{ .839,  .863,  .894}$\sim$4 FPS & \cellcolor[rgb]{ .839,  .863,  .894}\textcolor[rgb]{ 1,  0,  0}{\XSolidBrush} & \cellcolor[rgb]{ .839,  .863,  .894}\textcolor[rgb]{ 0,  .69,  .314}{\checkmark} & \cellcolor[rgb]{ .839,  .863,  .894}\textcolor[rgb]{ 0,  .69,  .314}{\checkmark} & \cellcolor[rgb]{ .839,  .863,  .894}\textcolor[rgb]{ 1,  0,  0}{\XSolidBrush} \\
\cmidrule{3-17}          &       & \cellcolor[rgb]{ .839,  .863,  .894}Kim \etal\cite{Kim2021Eventguided} & \cellcolor[rgb]{ .839,  .863,  .894}\textcolor[rgb]{ 1,  0,  0}{\XSolidBrush} & \cellcolor[rgb]{ .839,  .863,  .894}\textcolor[rgb]{ 0,  .69,  .314}{\checkmark} & \cellcolor[rgb]{ .839,  .863,  .894}\textcolor[rgb]{ 0,  .69,  .314}{\checkmark} & \cellcolor[rgb]{ .839,  .863,  .894}Frame+event sensor & \cellcolor[rgb]{ .839,  .863,  .894}Chip synchronization & \cellcolor[rgb]{ .839,  .863,  .894}DAVIS346 color & \cellcolor[rgb]{ .839,  .863,  .894}346$\times$260 & \cellcolor[rgb]{ .839,  .863,  .894}346$\times$260 & \cellcolor[rgb]{ .839,  .863,  .894}color & \cellcolor[rgb]{ .839,  .863,  .894}$\sim$3 FPS & \cellcolor[rgb]{ .839,  .863,  .894}\textcolor[rgb]{ 0,  .69,  .314}{\checkmark} & \cellcolor[rgb]{ .839,  .863,  .894}\textcolor[rgb]{ 1,  0,  0}{\XSolidBrush} & \cellcolor[rgb]{ .839,  .863,  .894}\textcolor[rgb]{ 1,  0,  0}{\XSolidBrush} & \cellcolor[rgb]{ .839,  .863,  .894}\textcolor[rgb]{ 1,  0,  0}{\XSolidBrush} \\
\cmidrule{2-17}          & \multicolumn{1}{c|}{\multirow{2}[4]{*}{\textbf{Real}}} &   REBlur\cite{Sun2021EventBasedFF} & \textcolor[rgb]{ 0,  .69,  .314}{\checkmark} & \textcolor[rgb]{ 0,  .69,  .314}{\checkmark} & \textcolor[rgb]{ 0,  .69,  .314}{\checkmark} & Repetitive motion scenes & Mark points matching & DAVIS346 mono & 346$\times$260 & 346$\times$260 & gray  & $\textless$ 50 FPS & \textcolor[rgb]{ 1,  0,  0}{\XSolidBrush} & \textcolor[rgb]{ 1,  0,  0}{\XSolidBrush} & \textcolor[rgb]{ 0,  .69,  .314}{\checkmark} & \textcolor[rgb]{ 1,  0,  0}{\XSolidBrush} \\
\cmidrule{3-17}          &       & \cellcolor[rgb]{ 1,  .953,  .792}\textbf{\textsc{EventAid-B}} & \cellcolor[rgb]{ 1,  .953,  .792}\textcolor[rgb]{ 0,  .69,  .314}{\checkmark} & \cellcolor[rgb]{ 1,  .953,  .792}\textcolor[rgb]{ 0,  .69,  .314}{\checkmark} & \cellcolor[rgb]{ 1,  .953,  .792}\textcolor[rgb]{ 0,  .69,  .314}{\checkmark} & \cellcolor[rgb]{ 1,  .953,  .792}Beam splitter & \cellcolor[rgb]{ 1,  .953,  .792}External clock triggering & \cellcolor[rgb]{ 1,  .953,  .792}Prophesee & \multicolumn{1}{c|}{\cellcolor[rgb]{ 1,  .953,  .792}$\sim$835$\times$620} & \multicolumn{1}{c|}{\cellcolor[rgb]{ 1,  .953,  .792}$\sim$835$\times$620} & \cellcolor[rgb]{ 1,  .953,  .792}color & \cellcolor[rgb]{ 1,  .953,  .792}30 FPS & \cellcolor[rgb]{ 1,  .953,  .792}\textcolor[rgb]{ 0,  .69,  .314}{\checkmark} & \cellcolor[rgb]{ 1,  .953,  .792}\textcolor[rgb]{ 0,  .69,  .314}{\checkmark} & \cellcolor[rgb]{ 1,  .953,  .792}\textcolor[rgb]{ 0,  .69,  .314}{\checkmark} & \cellcolor[rgb]{ 1,  .953,  .792}\textcolor[rgb]{ 0,  .69,  .314}{\checkmark} \\
    \midrule[2pt]
    \multicolumn{1}{c|}{\multirow{4}[8]{*}{\textbf{\makecell[c]{Event-aided\\image\\super resolution}}}} & \multicolumn{1}{c|}{\multirow{3}[6]{*}{\textbf{Simulation}}} & \cellcolor[rgb]{ .839,  .863,  .894}ESC\cite{E2SRI_pami} & \cellcolor[rgb]{ .839,  .863,  .894}\textcolor[rgb]{ 1,  0,  0}{\XSolidBrush} & \cellcolor[rgb]{ .839,  .863,  .894}\textcolor[rgb]{ 1,  0,  0}{\XSolidBrush} & \cellcolor[rgb]{ .839,  .863,  .894}\textcolor[rgb]{ 0,  .69,  .314}{\checkmark} & \cellcolor[rgb]{ .839,  .863,  .894}- & \cellcolor[rgb]{ .839,  .863,  .894}- & \cellcolor[rgb]{ .839,  .863,  .894}Simulation & \cellcolor[rgb]{ .839,  .863,  .894}128$\times$128 & \cellcolor[rgb]{ .839,  .863,  .894}512$\times$512 & \cellcolor[rgb]{ .839,  .863,  .894}gray & \cellcolor[rgb]{ .839,  .863,  .894}- & \cellcolor[rgb]{ .839,  .863,  .894}\textcolor[rgb]{ 0,  .69,  .314}{\checkmark} & \cellcolor[rgb]{ .839,  .863,  .894}\textcolor[rgb]{ 0,  .69,  .314}{\checkmark} & \cellcolor[rgb]{ .839,  .863,  .894}\textcolor[rgb]{ 0,  .69,  .314}{\checkmark} & \cellcolor[rgb]{ .839,  .863,  .894}\textcolor[rgb]{ 0,  .69,  .314}{\checkmark} \\
\cmidrule{3-17}          &       & \cellcolor[rgb]{ .839,  .863,  .894}GoPro+V2E\cite{Han-iccv21} & \cellcolor[rgb]{ .839,  .863,  .894}\textcolor[rgb]{ 1,  0,  0}{\XSolidBrush} & \cellcolor[rgb]{ .839,  .863,  .894}\textcolor[rgb]{ 1,  0,  0}{\XSolidBrush} & \cellcolor[rgb]{ .839,  .863,  .894}\textcolor[rgb]{ 0,  .69,  .314}{\checkmark} & \cellcolor[rgb]{ .839,  .863,  .894}- & \cellcolor[rgb]{ .839,  .863,  .894}- & \cellcolor[rgb]{ .839,  .863,  .894}Simulation & \cellcolor[rgb]{ .839,  .863,  .894}320$\times$180 & \cellcolor[rgb]{ .839,  .863,  .894}1280$\times$720 & \cellcolor[rgb]{ .839,  .863,  .894}color & \cellcolor[rgb]{ .839,  .863,  .894}- & \cellcolor[rgb]{ .839,  .863,  .894}\textcolor[rgb]{ 0,  .69,  .314}{\checkmark} & \cellcolor[rgb]{ .839,  .863,  .894}\textcolor[rgb]{ 0,  .69,  .314}{\checkmark} & \cellcolor[rgb]{ .839,  .863,  .894}\textcolor[rgb]{ 0,  .69,  .314}{\checkmark} & \cellcolor[rgb]{ .839,  .863,  .894}\textcolor[rgb]{ 0,  .69,  .314}{\checkmark} \\
\cmidrule{3-17}          &       & \cellcolor[rgb]{ .839,  .863,  .894}GoPro+ESIM\cite{esl-eccv20} & \cellcolor[rgb]{ .839,  .863,  .894}\textcolor[rgb]{ 1,  0,  0}{\XSolidBrush} & \cellcolor[rgb]{ .839,  .863,  .894}\textcolor[rgb]{ 1,  0,  0}{\XSolidBrush} & \cellcolor[rgb]{ .839,  .863,  .894}\textcolor[rgb]{ 0,  .69,  .314}{\checkmark} & \cellcolor[rgb]{ .839,  .863,  .894}- & \cellcolor[rgb]{ .839,  .863,  .894}- & \cellcolor[rgb]{ .839,  .863,  .894}Simulation & \cellcolor[rgb]{ .839,  .863,  .894}320$\times$180 & \cellcolor[rgb]{ .839,  .863,  .894}1280$\times$720 & \cellcolor[rgb]{ .839,  .863,  .894}gray & \cellcolor[rgb]{ .839,  .863,  .894}- & \cellcolor[rgb]{ .839,  .863,  .894}\textcolor[rgb]{ 0,  .69,  .314}{\checkmark} & \cellcolor[rgb]{ .839,  .863,  .894}\textcolor[rgb]{ 0,  .69,  .314}{\checkmark} & \cellcolor[rgb]{ .839,  .863,  .894}\textcolor[rgb]{ 0,  .69,  .314}{\checkmark} & \cellcolor[rgb]{ .839,  .863,  .894}\textcolor[rgb]{ 0,  .69,  .314}{\checkmark} \\
\cmidrule{2-17}          & \multicolumn{1}{c|}{\textbf{Real}} & \cellcolor[rgb]{ 1,  .953,  .792}\textbf{\textsc{EventAid-S}} & \cellcolor[rgb]{ 1,  .953,  .792}\textcolor[rgb]{ 1,  0,  0}{\XSolidBrush} & \cellcolor[rgb]{ 1,  .953,  .792}\textcolor[rgb]{ 0,  .69,  .314}{\checkmark} & \cellcolor[rgb]{ 1,  .953,  .792}\textcolor[rgb]{ 0,  .69,  .314}{\checkmark} & \cellcolor[rgb]{ 1,  .953,  .792}Beam splitter & \cellcolor[rgb]{ 1,  .953,  .792}External clock triggering & \cellcolor[rgb]{ 1,  .953,  .792}Prophesee & \multicolumn{1}{c|}{\cellcolor[rgb]{ 1,  .953,  .792}1270$\times$710} & \multicolumn{1}{c|}{\cellcolor[rgb]{ 1,  .953,  .792}2540$\times$1420} & \cellcolor[rgb]{ 1,  .953,  .792}color & \cellcolor[rgb]{ 1,  .953,  .792}30 FPS & \cellcolor[rgb]{ 1,  .953,  .792}\textcolor[rgb]{ 0,  .69,  .314}{\checkmark} & \cellcolor[rgb]{ 1,  .953,  .792}\textcolor[rgb]{ 0,  .69,  .314}{\checkmark} & \cellcolor[rgb]{ 1,  .953,  .792}\textcolor[rgb]{ 0,  .69,  .314}{\checkmark} & \cellcolor[rgb]{ 1,  .953,  .792}\textcolor[rgb]{ 0,  .69,  .314}{\checkmark} \\
    \midrule[2pt]
    \multicolumn{1}{c|}{\multirow{5}[10]{*}{\textbf{\makecell[c]{Event-aided\\high dynamic range\\image reconstruction}}}} & \multicolumn{1}{c|}{\textbf{Simulation}} & \cellcolor[rgb]{ .839,  .863,  .894}Yang \etal\cite{Yang_2023_HDR} & \cellcolor[rgb]{ .839,  .863,  .894}\textcolor[rgb]{ 1,  0,  0}{\XSolidBrush} & \cellcolor[rgb]{ .839,  .863,  .894}\textcolor[rgb]{ 1,  0,  0}{\XSolidBrush} & \cellcolor[rgb]{ .839,  .863,  .894}\textcolor[rgb]{ 0,  .69,  .314}{\checkmark} & \cellcolor[rgb]{ .839,  .863,  .894}- & \cellcolor[rgb]{ .839,  .863,  .894}- & \cellcolor[rgb]{ .839,  .863,  .894}Simulation & \multicolumn{1}{c|}{\cellcolor[rgb]{ .839,  .863,  .894}256$\times$256} & \multicolumn{1}{p{5.725em}|}{\cellcolor[rgb]{ .839,  .863,  .894}\quad256$\times$256} & \cellcolor[rgb]{ .839,  .863,  .894}color & \cellcolor[rgb]{ .839,  .863,  .894}- & \cellcolor[rgb]{ .839,  .863,  .894}\textcolor[rgb]{ 0,  .69,  .314}{\checkmark} & \cellcolor[rgb]{ .839,  .863,  .894}\textcolor[rgb]{ 1,  0,  0}{\XSolidBrush} & \cellcolor[rgb]{ .839,  .863,  .894}\textcolor[rgb]{ 1,  0,  0}{\XSolidBrush} & \cellcolor[rgb]{ .839,  .863,  .894}\textcolor[rgb]{ 0,  .69,  .314}{\checkmark} \\
\cmidrule{2-17}          & \multicolumn{1}{c|}{\multirow{4}[8]{*}{\textbf{Real}}} &   Han \etal\cite{han2020hdr} & \textcolor[rgb]{ 0,  .69,  .314}{\checkmark} & \textcolor[rgb]{ 0,  .69,  .314}{\checkmark} & \textcolor[rgb]{ 1,  0,  0}{\XSolidBrush} & Beam splitter & Mark points matching & DAVIS240 & 240$\times$180 & 1520$\times$1440 & color & 20 FPS & \textcolor[rgb]{ 0,  .69,  .314}{\checkmark} & \textcolor[rgb]{ 0,  .69,  .314}{\checkmark} & \textcolor[rgb]{ 1,  0,  0}{\XSolidBrush} & \textcolor[rgb]{ 1,  0,  0}{\XSolidBrush} \\
\cmidrule{3-17}          &       &   HES-HDR\cite{Han_HDR_pami} & \textcolor[rgb]{ 0,  .69,  .314}{\checkmark} & \textcolor[rgb]{ 0,  .69,  .314}{\checkmark} & \textcolor[rgb]{ 1,  0,  0}{\XSolidBrush} & Beam splitter & Mark points matching & DAVIS346 mono & 329$\times$237 & 2032$\times$1446 & color & 20 FPS & \textcolor[rgb]{ 0,  .69,  .314}{\checkmark} & \textcolor[rgb]{ 0,  .69,  .314}{\checkmark} & \textcolor[rgb]{ 1,  0,  0}{\XSolidBrush} & \textcolor[rgb]{ 0,  .69,  .314}{\checkmark} \\
\cmidrule{3-17}          &       &   Yang \etal\cite{Yang_2023_HDR} & \textcolor[rgb]{ 0,  .69,  .314}{\checkmark} & \textcolor[rgb]{ 0,  .69,  .314}{\checkmark} & \textcolor[rgb]{ 1,  0,  0}{\XSolidBrush} & Beam splitter & Mark points matching & DAVIS346 color & 346$\times$260 & 346$\times$260 & color &$\sim$10 FPS & \textcolor[rgb]{ 1,  0,  0}{\XSolidBrush} & \textcolor[rgb]{ 1,  0,  0}{\XSolidBrush} & \textcolor[rgb]{ 1,  0,  0}{\XSolidBrush} & \textcolor[rgb]{ 0,  .69,  .314}{\checkmark} \\
\cmidrule{3-17}          &       & \cellcolor[rgb]{ 1,  .953,  .792}\textbf{\textsc{EventAid-D}} & \cellcolor[rgb]{ 1,  .953,  .792}\textcolor[rgb]{ 0,  .69,  .314}{\checkmark} & \cellcolor[rgb]{ 1,  .953,  .792}\textcolor[rgb]{ 0,  .69,  .314}{\checkmark} & \cellcolor[rgb]{ 1,  .953,  .792}Reference & \cellcolor[rgb]{ 1,  .953,  .792}Beam splitter & \cellcolor[rgb]{ 1,  .953,  .792}External clock triggering & \cellcolor[rgb]{ 1,  .953,  .792}Prophesee & \cellcolor[rgb]{ 1,  .953,  .792}$\sim$800$\times$500 & \cellcolor[rgb]{ 1,  .953,  .792}$\sim$800$\times$500 & \cellcolor[rgb]{ 1,  .953,  .792}color & \cellcolor[rgb]{ 1,  .953,  .792}30 FPS & \cellcolor[rgb]{ 1,  .953,  .792}\textcolor[rgb]{ 0,  .69,  .314}{\checkmark} & \cellcolor[rgb]{ 1,  .953,  .792}\textcolor[rgb]{ 0,  .69,  .314}{\checkmark} & \cellcolor[rgb]{ 1,  .953,  .792}\textcolor[rgb]{ 0,  .69,  .314}{\checkmark} & \cellcolor[rgb]{ 1,  .953,  .792}\textcolor[rgb]{ 0,  .69,  .314}{\checkmark} \\
    \bottomrule[3pt]
    \end{tabular}}}%
  \label{tab:summary}%
\end{table*}%

\customparagraph{Event-based video reconstruction.} The basic task of bridging events and images that directly reconstructs images from pure event signals, which can be formulated as a process of $E_{t_0:t_1}\rightarrow I_{t_1}$, where the $E_{t_0:t_1}$ denotes the event stream triggered between $t_0$ and $t_1$. This is an ill-posed problem because the event signals only record the intensity change but not the absolute intensity in the scene, so it is difficult to accurately measure the light intensity via events. To solve this problem, Barua \etal \cite{barua2016direct, reinbacher2016realtime, Zhang22pami} propose to use the optical flow consistency hypothesis and motion compensation to obtain the gradient of images from events, and then employ the Poisson reconstruction method to restore the image, \ie, $I_{t_1}=\mathcal{F}_\text{passion}(\mathcal{F}_\text{warp}(E_{t_0:t_1}))$ (\cf{, Eq.~\textcolor{red}{1} in \fref{fig:overview}). Deep learning-based methods \cite{e2vid-cvpr19, e2vid-pami19, ced, scheerlinck2020fast, SPADE-E2VID} directly learn the mapping model from events to images by $I_{t_1}=\mathcal{F}_\text{rec}(E_{t_0:t_1})$ (\cf{, Eq.~\textcolor{red}{2}). It is worth noting that this task cannot reconstruct textures in the static scene.

\customparagraph{Event-aided HFR video reconstruction.} This task aims to interpolate new frames, \ie{, reconstruct latent frames}, between two adjacent frames with the assistance of events, which is formulated as $I_{t_0}\&E_{t_0:t_1}\rightarrow I_{t_1}$. Since events record the logarithmic changes of $I_{t_0}$ over $t_0:t_1$ with high time accuracy, the $I_{t_1}$ in the logarithmic domain (\ie, $L_{t_1}$) can be easily obtained by the events synthesis model $L_{t_1}=L_{t_0}+c\cdot\int E_{t_0:t_1}$ (\cf{, Eq.~\textcolor{red}{3}), despite the interference of event noise \cite{Tobi14ISCAS, wang2021asynchronous}. In order to improve performance, deep learning-based methods generally employ the events synthesis model to constraint intensity values and event-based optical flow estimation to constraint motion trajectories in reconstructed videos, then use a fusion model to fuse the two branches and output the final result \cite{Timelens, timelens++, SuperFast, Paikin_2021_CVPRW}, formulated as $I_{t_1}=\mathcal{F}_\text{fusion}(\mathcal{F}_\text{syn}(I_{t_0}, E_{t_0:t_1}),\mathcal{F}_\text{warp}(I_{t_0}, E_{t_0:t_1}))$ (\cf{, Eq.~\textcolor{red}{4}).

\customparagraph{Event-aided image deblurring.} This task aims to restore a clear image from the long-exposure image suffering from motion blur, formulated as $I^\text{blur}\&E_{t_0:t_1}\rightarrow I^\text{clear}$, where $t_0:t_1$ corresponds to the exposure time period. Pan \etal \cite{edi-cvpr19} find the event-based double integral model to bridge $I^\text{blur}$ and $I^\text{clear}$ via events and reconstruct the clear image via $L^\text{clear} = L^\text{blur} - c\cdot \iint E_{t_0:t_1}$ (\cf{, Eq.~\textcolor{red}{5}). Learning-based methods \cite{sun2023deblur, chen22tmm, Sun2021EventBasedFF, lin2020ledvdi} continuously improve the deblurring performance by upgrading the network model. Optical estimation is also introduced to improve performance \cite{jiang2020learning}, \ie{}, $I^\text{clear}=\mathcal{F}_\text{fusion}(\mathcal{F}_\text{syn}(I^\text{blur}, E_{t_0:t_1}),\mathcal{F}_\text{warp}(I^\text{blur}, E_{t_0:t_1}))$ (\cf{, Eq.~\textcolor{red}{6}). Due to the high temporal resolution of events, most methods achieve intra-frame interpolation as well.

\customparagraph{Event-aided image super-resolution.} This task aims to reconstruct a high-resolution image from a low-resolution image by converting event-recorded motion information into sub-pixel shifts, \ie{}, $I^{\text{LR}}_{t_0}\&E_{t_0-w:t_0+w}\rightarrow I^{\text{SR}}_{t_0}$, where $w$ adjust time window length. EvIntSR \cite{Han-iccv21} and E2SRI \cite{e2sri-cvpr20, E2SRI_pami} generally convert event data to multiple latent intensity frames and learn to mix the frame sequence to achieve super-resolution, expressed as $I^{\text{SR}}_{t_0} = \mathcal{F}_\text{mix}(I^{\text{LR}}_{t_0}, \mathcal{F}_\text{rec}(E_{t_0-w:t_0-w+\epsilon}), \dots, \mathcal{F}_\text{rec}(E_{t_0+w-\epsilon:t_0+w}))$ (\cf{, Eq.~\textcolor{red}{7}), $\epsilon$ is the time length of events to convert each latent frames. EventSR \cite{eventsr-cvpr20} can also achieve image SR, while it mainly learns the mapping from LR images generated by events to HR images through GAN-based methods \cite{wang2018esrgan}.

\customparagraph{Event-aided HDR image reconstruction.} This task aims to recover an HDR image from a low dynamic range (LDR) image by extracting texture features of over-/under-exposed areas from events in dynamic scenes, \ie{}, $I^\text{LDR}_{t_1}\&E_{t_0:t_1}\rightarrow I^\text{HDR}_{t_1}$. Han \etal \cite{han2020hdr, Han_HDR_pami} first explore this task and propose to reconstruct an intensity frame from events before fusing it with input LDR images via a refinement network module, expressed as $I^\text{HDR}_{t_1} = \mathcal{F}_\text{fusion}(I^\text{LDR}_{t_1}, \mathcal{F}_\text{rec}(E_{t_0:t_1}))$ (\cf{, Eq.~\textcolor{red}{8}). Yang \etal \cite{Yang_2023_HDR} eliminate the step of reconstructing the image from events and recover an HDR image by $I^\text{HDR}_{t_1} = \mathcal{F}_\text{fusion}(I^\text{LDR}_{t_1}, E_{t_0:t_1})$ (\cf{, Eq.~\textcolor{red}{9}).

\subsection{Evaluation datasets for event-aided imaging} 

The field of conventional image/video enhancement already has a large amount of research on benchmark evaluation. Such as Köhler \etal \cite{SR_real_bench} propose a real-captured dataset to benchmark SR and Rim \etal \cite{rim_2020_ECCV} propose a real-captured dataset to benchmark deblurring. In contrast, due to the lack of comprehensive real-captured datasets and quantitative benchmarks, the performance of event-aided image/video enhancement methods on real data is still largely unexplored. In \Tref{tab:summary}, we summarize widely used evaluation datasets for five event-aided image/video enhancement tasks to compare their characteristics. The proposed real-captured dataset \textsc{EventAid} is also added to this comparison to highlight its advantage.

We focus on the following properties to evaluate the characteristics of these datasets: (1) Whether the triplet (input frames, input events, and ground truth) are real-captured data: Compared to using simulated data as evaluation datasets, real-captured data enables benchmarking the enhancement performance of the algorithm in real-world scenarios. However, since it is extremely challenging to collect real-captured data simultaneously, existing datasets often complete the triplet data by simulating events or synthesizing blur images, LDR images, \etc, which will introduce a real-sim gap and make the benchmark results less convincing. (2) Spatiotemporal synchronization manner of event sensor and frame-based sensor\footnote{The illustrations of different spatiotemporal synchronization ways are shown in Sec. \textcolor{red}{0} of the supplementary material.}: To capture real data, there are four spatial matching patterns, where ``frame+event sensor'' is the best solution for synchronization but lacks the flexibility of switching cameras, ``dual camera setup'' can easily replace different frame cameras but disparity prevents pixels from being accurately aligned, ``repetitive motion scenes'' is difficult to collect diverse scenes. In contrast, ``beam splitter'' can effectively avoid the above problems. There are three temporal synchronization patterns, where ``chip synchronization'' is the best choice, and ``external clock triggering'' can also achieve microsecond level synchronization error. (3) Performance parameters of event sensor and frame-based sensor: Higher resolution and frame rates help to collect high-quality data. (4) The diversity of the captured scenes: Datasets covering diverse scenarios help evaluate the robustness of algorithms.

\begin{figure*}[t]
    \centering
    \includegraphics[width=1\linewidth]{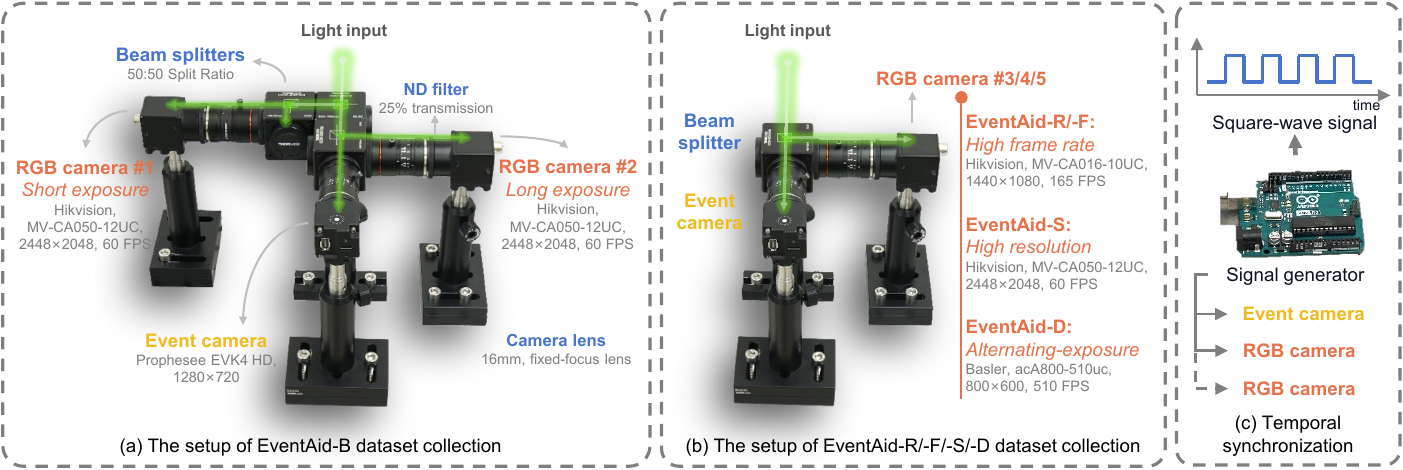}
    \caption{The equipment setup we used to collect the proposed dataset. (a) For the \textsc{EventAid-B} dataset, we use one RGB camera to capture long-exposure blur images as the input, and another RGB camera to capture short-exposure clear images as the ground truth, the corresponding events are captured by one event camera. Three $50:50$ split ratio beam splitters are docked and mounted in front of the lenses. (b) For the \textsc{EventAid-R/-F/-S/-D} dataset, we collocate an event camera and an RGB camera by mounting a $50:50$ split ratio beam splitter in front of them. For each task, an RGB camera with corresponding attributes is selected to ensure that effective ground truths are captured. (c) We use a signal generator to simultaneously send square-wave signals to all cameras to achieve synchronized shooting.}
    \label{fig:setup}
    \vspace{-10pt}
\end{figure*}

\begin{figure}[t]
    \centering
    \includegraphics[width=1\linewidth]{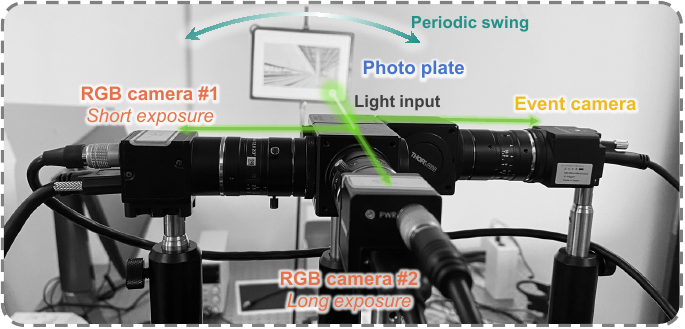}
    \caption{The equipment setup we used for the controlled experiment of event-aided image deblurring task.}
    \label{fig:control}
    \vspace{-15pt}
\end{figure}

\customparagraph{Event-based video reconstruction.} For this task, evaluation data should contain input events and ground truth frames. EventNFS \cite{EventZoom} develops a display-camera system to observe real-scenario event data via playback of $240FPS$ 720p videos on the display, while the real-sim gap of events still exists due to the relatively low refresh rate and dynamic range display. IJRR \cite{davis}, HQF \cite{sim-to-real-gap-eccv20}, DVS-Dark \cite{ev-sid-eccv20}, MVSEC \cite{mvsec-ral18} and CED \cite{ced} all capture event data with the DAVIS series cameras \cite{davis240, davis346} and use APS as the ground truth frames. However, these cameras have low resolution and frame rate, and APS suffers from severe noise. In contrast, \textsc{EventAid-R} collects real-captured data at an average $954\times636$ resolution and $150FPS$, enabling the benchmark results to meet the requirements of real-world application scenarios.

\customparagraph{Event-aided HFR video reconstruction.} For this task, evaluation data should contain the triplet: input events, input LFR frame sequence, and ground truth HFR frames. The simulated datasets \cite{tulyakov2019learning, esl-eccv20} take in HFR video datasets as ground truth and pass them into simulators to generate event signals. SloMo-DVS \cite{Yu_2021_ICCV} and GEF \cite{gef-cvpr20, gef-tpami}, while real-captured, also suffer from low resolution and low frame rates. HS-ERGB \cite{Timelens} first collects HFR and HR videos as the ground truth, while the baseline existing in the dual camera system introduces inevitable errors. BS-ERGB \cite{timelens++} selects an LFR frame-based camera and the two cameras mount different lenses. The proposed \textsc{EventAid-F} avoids the above shortcomings and provides ground truth videos of $150FPS$ for the algorithms to be evaluated.

\customparagraph{Event-aided image deblurring.} This task requires evaluation data containing the triplet: input events, input blur images, and ground truth blur-free images. Different from the above two tasks, simultaneous capturing of blur and blur-free images greatly increases the difficulty of data collection. Therefore, most datasets simulate blurry images by averaging multi-frames \cite{esl-eccv20, jiang2020learning, ev_deblur, Kim2021Eventguided}. To achieve real-captured data collection, REBlur \cite{Sun2021EventBasedFF} performs controlled experiments indoors to collect the triplet data by repeating the same motion scenario multiple times, which can only capture indoor scenes with nondiversity. To collect diversity and real-world datasets, the proposed \textsc{EventAid-B} first captures all real-captured triplet data by synchronizing two frame cameras and an event camera via beam splitters.

\customparagraph{Event-aided image super-resolution.} This task requires evaluation data containing input events, input LR images, and ground truth HR images. Since this task is still in its initial exploration stage, the existing evaluation datasets are all simulated. The proposed \textsc{EventAid-S} dataset is the first real-captured evaluation dataset, which captures $2\times$ HR frames as the ground truth and $1\times$ events as the LR inputs. The input $1\times$ LR frames are downsampled from HR ones following the process in single image SR tasks.

\customparagraph{Event-aided HDR image reconstruction.} For this task, evaluation data contains input events, input LDR images, and ground truth HDR images. Existing datasets, \ie{}, Han \etal \cite{han2020hdr}, HES-HDR \cite{Han_HDR_pami}, and Yang \etal \cite{Yang_2023_HDR} only contain input events and LDR images. They mainly evaluate the quality of the reconstructed HDR images through no-reference quality assessment. \textsc{EventAid-D} uses an alternating-exposure camera to cyclically get short-/middle-/long-exposure LDR images as the diverse input data, and mix multi-exposure images to restore HDR images as the reference.

\begin{figure*}[t]
    \centering
    \includegraphics[width=1\linewidth]{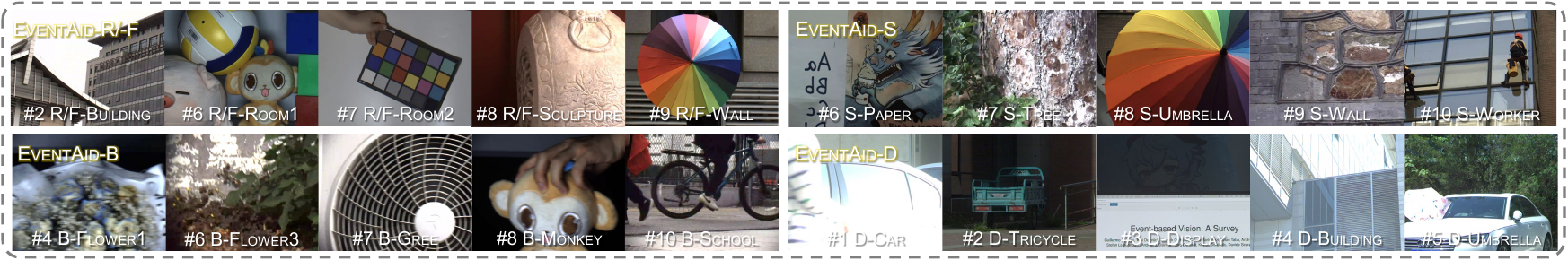}
    \caption{Overview of the real-world scenes on the proposed \textsc{EventAid-R/-F/-S/-B/-D} datasets. The corresponding number and name of each scene are marked below the image. The scenes include indoor, outdoor, global and local motion, slow and fast motion, over and under exposure, and high texture scenes.}
    \label{fig:dataset}

\end{figure*}

\section{\textsc{EventAid} Dataset collecting}

This section introduces the collection process of the proposed \textsc{EventAid} Dataset. \Fref{fig:setup} shows the equipment setup we used to collect datasets.

\subsection{Sensor and optics configuration}

To collect datasets with high imaging quality, spatiotemporal synchronization, and unified scale, we use one Prophesee EVK4 HD ($1280\times720$) event camera to capture event signals, two Hikvision MV-CA050-12UC RGB cameras ($2448\times2048$, $60FPS$) to simultaneously capture short-exposure and long-exposure frames for deblurring task, one Hikvision MV-CA050-12UC RGB cameras ($2448\times2048$, $60FPS$) to capture HR images, one Hikvision MV-CA016-10UC RGB camera ($1440\times1080$, $165FPS$) to capture HFR frames, and one Basler acA800-510uc RGB camera ($800\times600$, $510FPS$) to capture alternating-exposure frame for HDR reconstruction task.
For all cameras, we mount the same lenses ($16mm$, $\text{F}=1:2.8$, C-mount, fixed focus) to avoid the influence of focal length and distortion differences. During the capturing process, we balance image quality and depth of field to determine aperture parameters and keep them consistent across all lenses. For scenes with multiple objects at different depths, we adjust focus rings to ensure the objects at the image center are in focus. We use Thorlabs CCM1-BS013 beam splitters ($50:50$ Split Ratio) to share light input for multiple cameras. In addition, when collecting \textsc{EventAid-B}, we set a 25\% transmission ND filter to ensure luminosity consistency for short-exposure and long-exposure cameras.

Since event-aided image deblurring has been widely studied, and the algorithm performance is related to the degree of blur caused by motion, we execute a controllable experiment to evaluate the limits of blur levels that existing methods can withstand. \Fref{fig:control} shows the equipment setup and the experiment site layout of the controlled experiment. We use a servo steering gear to control a rigid rod to swing periodically at an opening angle of 120\degree, and fix a flat plate with a high-definition photo $1m$ away from the center of rotation as the main shooting target. The equipment setup is placed about $1m$ in front of the photo. Thus, when the exposure time of the cameras is fixed, we can adjust the motion speed of the photo by adjusting the swing period of the steering gear to obtain frames with different blur degrees. We set the exposure time of the short exposure camera to $2ms$ and the long exposure camera to $8ms$. The swing period is sequentially sampled at intervals of $0.25s$ from $1.5s$ to $4s$. We set up a DC fill light behind the scenes to ensure clear frames are less affected by noise imaging.

\begin{figure*}[t]
    \centering
    \includegraphics[width=.96\linewidth]{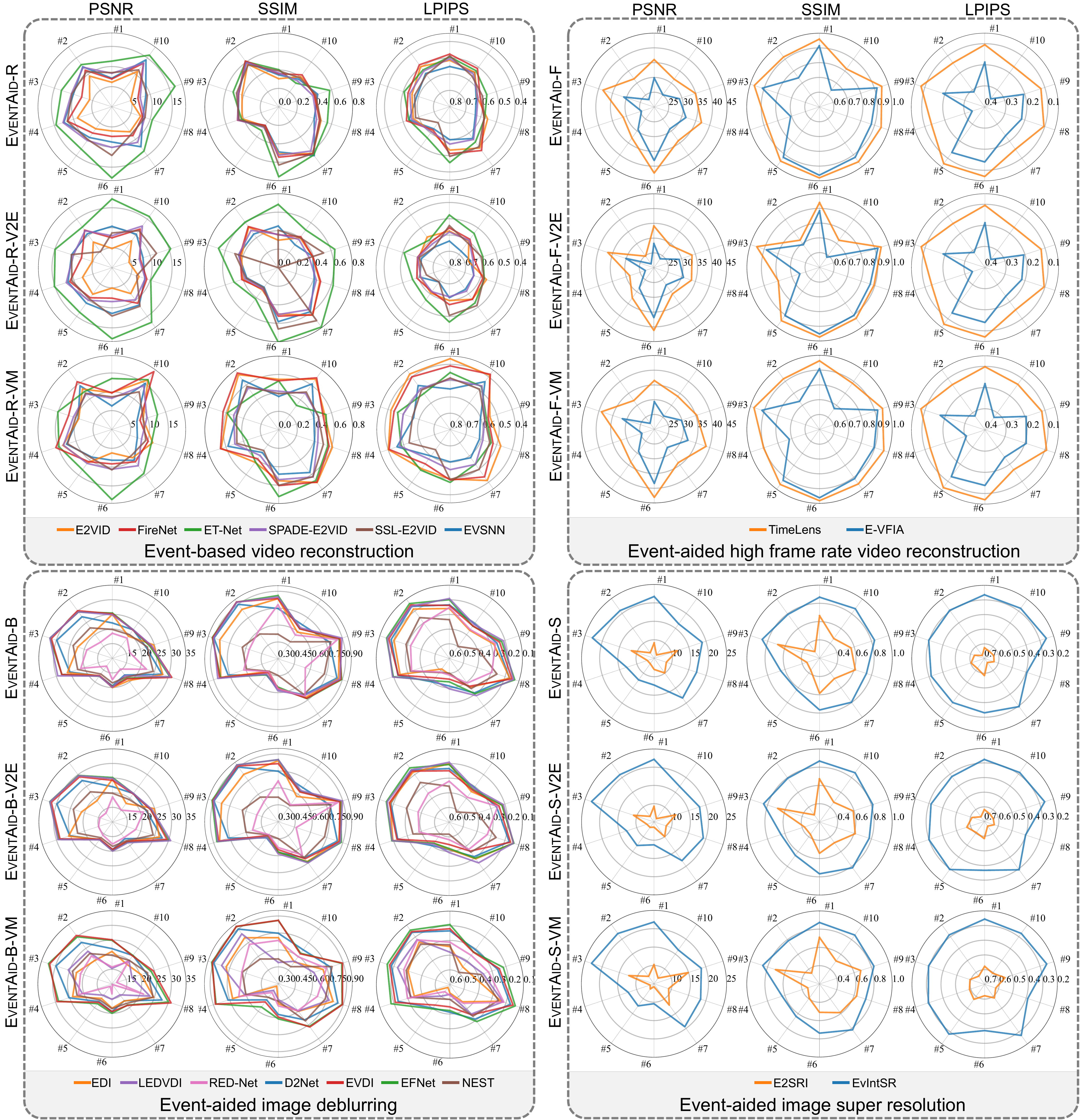}
    \caption{The benchmark results for four tasks on real-captured \textsc{EventAid} and simulated \textsc{EventAid-V2E/-VM} dataset. In the radar plots, the farther away from the center of the circle, the better the performance. The 10 axes in each radar plot correspond to the 10 groups in each of the dataset \textsc{EventAid-R/-F/-B/-S}.}
    \label{fig:radar}
    \vspace{-10pt}
\end{figure*}

\subsection{Spatial matching}

Since the deblurring task requires the simultaneous collection of blurry frames, blur-free frames, and event signals, it requires a hybrid of three cameras, while other tasks only require matching two cameras. Therefore, we use two types of camera hybrid system setups to collect data, as shown in \fref{fig:setup} (a) and (b). The first setup used to collect the \textsc{EventAid-B} dataset contains three cameras and we use three $50:50$ split ratio beam splitters docked and mounted in front of the lenses to ensure the input light is evenly split across all three cameras, \ie{}, each camera receives an equal 25\% of the input light. The three beam splitters are tightly connected through two SM1 external thread couplers, each lens is tightly connected with a beam splitter through a M27-to-SM1 thread adapter. The interfaces on the beam splitter that are not connected to the lens are blocked by light shields to avoid light interference. All cameras are fixed to a breadboard via poles to ensure they remain stable during severe shaking. Although the above-mentioned tight connection can make the fields of view of the three cameras well overlapped, it is still difficult to avoid pixel-level misalignment, so it is necessary to fine-tune before we collect the dataset. Similar to Wang \etal \cite{gef-cvpr20}, we use a 13.9” $60Hz$ monitor for an offline geometric register for three signals. We consider the homography transform among three camera views. To extract key points from event data, we display a blinking checkerboard pattern on the monitor and integrate the captured events to form a checkerboard image. Then we use the calculated homography matrix to transfer the views of two RGB cameras to the view of the event camera to ensure that the three camera views are consistent.
The second setup used to collect the \textsc{EventAid-R/-F/-S/-D} contains two cameras and we use a $50:50$ split ratio beam splitter to connect them.

\subsection{Temporal synchronization}
We use an Arduino Uno Rev3 microcontroller board as the signal generator to simultaneously send 5-volt square-wave signals to all cameras for achieving synchronized capturing. The corresponding interface of the signal generator and the GPIO ports of the cameras are connected through cables. The RGB camera takes a frame on each rising edge of the square-wave signal. The Prophesee event camera starts to trigger events when it receives the first rising edge and marks each rising edge and falling edge with special marking events. By shooting a high-precision LED timer, we have confirmed that the time error after temporal synchronization is within $10ms$, which meets the time accuracy requirements of the tasks benchmarked in this paper. \textsc{EventAid-R} and \textsc{EventAid-F} set the frame rate of RGB cameras to $150FPS$, and the others to $30FPS$. When shooting the \textsc{EventAid-B} dataset, the long-exposure time is fixed to $4\times$ of the short-exposure time, which corresponds to the 25\% transmission of the ND filter. To collect images with different blur levels, the long exposure time is set in the range of $6ms$ to $20ms$.

\subsection{Scene diversity and dataset size}
We consider the scenario diversity of all sub-datasets. As shown in \fref{fig:dataset}, all sub-datasets include indoor and outdoor, global and local motion, slow and fast motion, and high texture scenes\footnote{More scenes, as well as completed scene numbers and names are recorded in the supplementary material.}. To analyze the stability for challenging scenes, we collect fast, non-linear motion, and smooth texture scenes in \textsc{EventAid-R/-F/-B}, text and complex textures scenes in \textsc{EventAid-S}, and wide dynamic range scenes in \textsc{EventAid-D}. 
We collect a large amount of test data in each sub-dataset, with \textsc{EventAid-R} containing 10 groups totaling 58,264 frames and 405 seconds, \textsc{EventAid-B} containing 10 groups totaling 7,436 frames and 247 seconds, \textsc{EventAid-S} containing 10 groups totaling 11,265 frames and 375 seconds, and \textsc{EventAid-D} containing 5 groups totaling 4,353 frames and 107 seconds. \textsc{EventAid-R} and \textsc{EventAid-F} share the same data.

We also collect two simulated datasets, \ie{}, \textsc{EventAid-V2E/-VM}, to evaluate and compare the real-sim gap of two widely used event simulators, \ie{}, V2E \cite{V2E} and DVS-Voltmeter \cite{DVS-voltmeter} on event-based video reconstruction, event-aided HFR video reconstruction, image deblurring, and image SR reconstruction tasks. To simulate the corresponding events, the ground truth videos of \textsc{EventAid-R/-F/-B/-S} are input into two simulators and generate event data. All parameters are set according to the author's suggestions.

\begin{figure*}[t]
    \centering
    \includegraphics[width=1\linewidth]{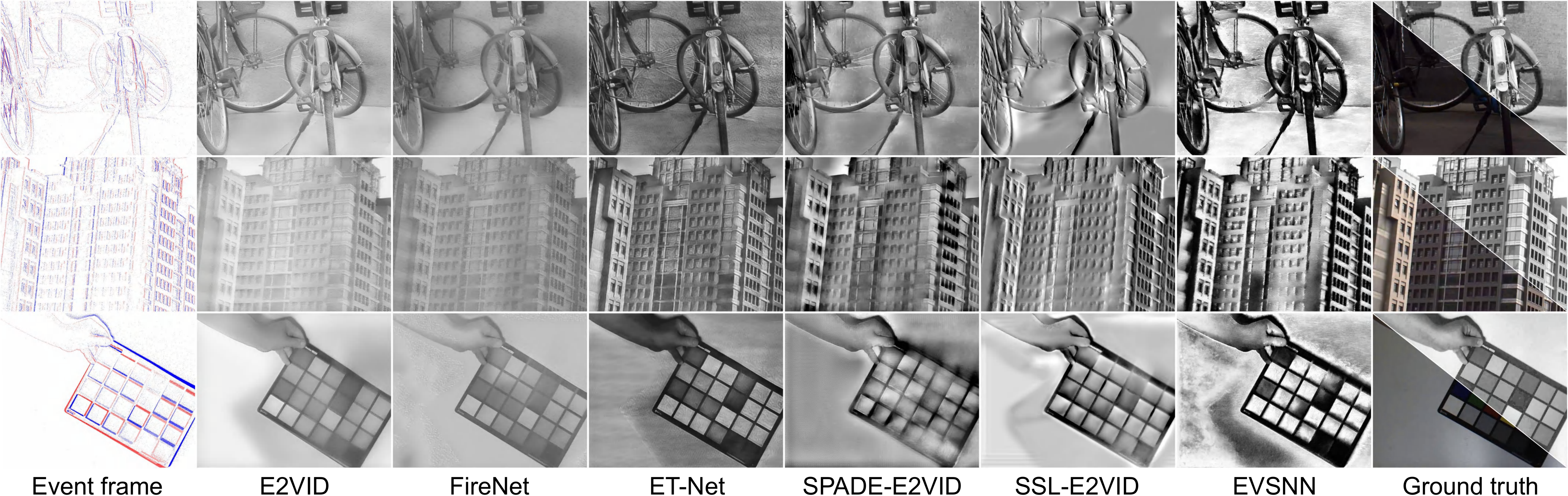}
    \caption{Event-based video reconstruction result examples from the \textsc{EventAid-R} dataset. We show the results output from E2VID \cite{e2vid-cvpr19, e2vid-pami19}, FireNet \cite{FireNet}, ET-Net \cite{etnet}, SPADE-E2VID \cite{SPADE-E2VID}, SSL-E2VID \cite{ssl-e2vid}, and EVSNN \cite{EVSNN}. In each square of the ground truth column, lower-left shows the RGB images, and upper-right shows the corresponding gray channel to facilitate comparison with the grayscale images produced from evaluated methods.}
    \label{fig:recons}
    \vspace{-10pt}
\end{figure*}

\begin{figure}[t]
    \centering
    \includegraphics[width=1\linewidth]{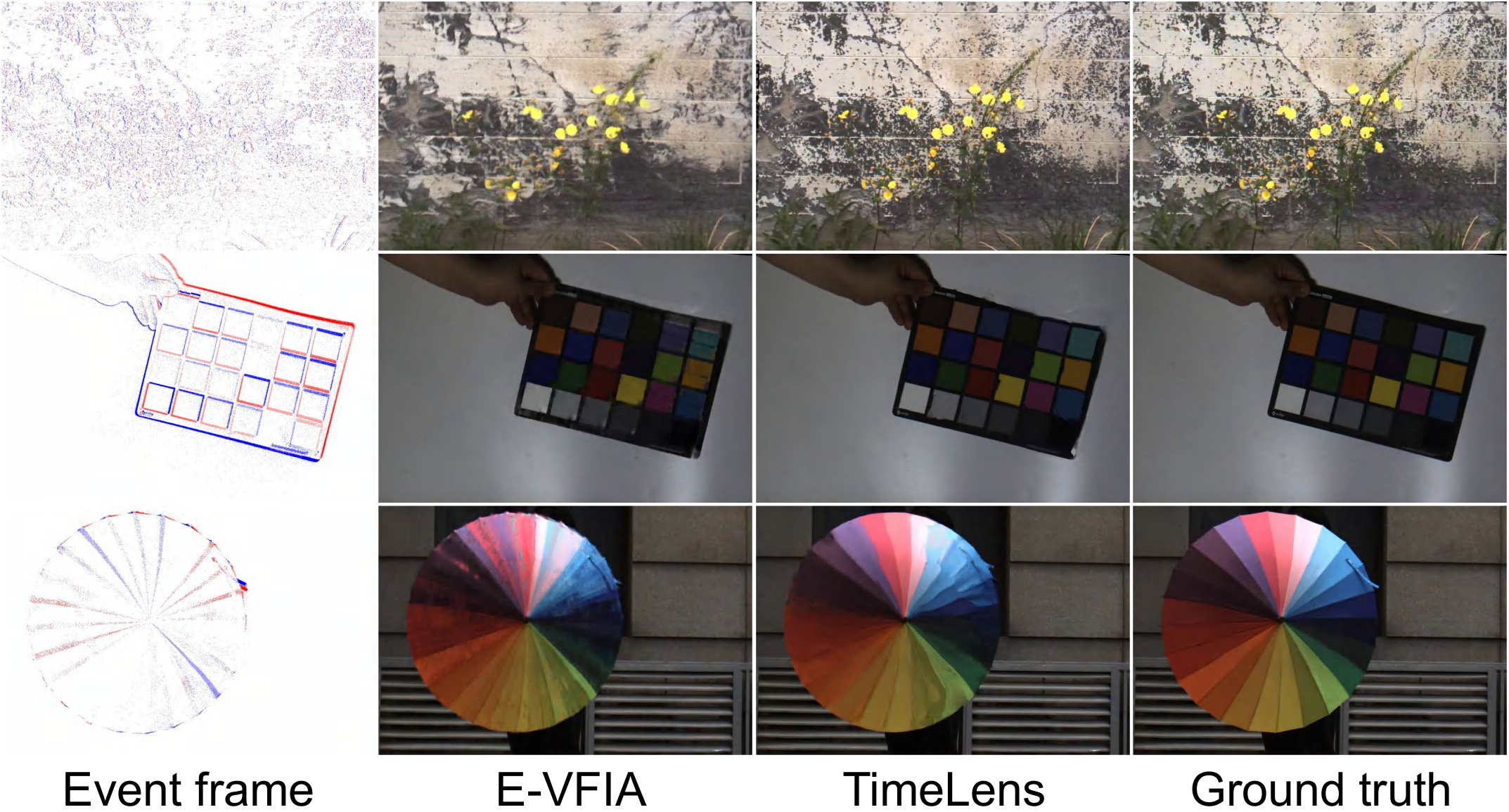}
    \caption{Event-aided HFR video reconstruction result examples from the \textsc{EventAid-F} dataset. We show the results produced from TimeLens \cite{Timelens} and E-VFIA \cite{E-VFIA}.}
    \label{fig:hfr}
    \vspace{-15pt}
\end{figure}

\section{Experiment and benchmark analysis}
This section showcases the benchmark results for five event-aided tasks on \textsc{EventAid} dataset.

\subsection{Methods and evaluation metrics}
We use the sub-datasets of \textsc{EventAid} to evaluate representative methods of five event-aided image/video enhancement tasks respectively. 
(1) For event-based video reconstruction task, we choose E2VID (\texttt{CVPR19}\cite{e2vid-cvpr19}, \texttt{TPAMI20}\cite{e2vid-pami19}), FireNet (\texttt{WACV20}\cite{FireNet}), ET-Net (\texttt{ICCV21}\cite{etnet}), SPADE-E2VID (\texttt{TIP21}\cite{SPADE-E2VID} ), SSL-E2VID (\texttt{ICCV21}\cite{ssl-e2vid}), and EVSNN (\texttt{CVPR22}\cite{EVSNN}). 
(2) For the event-aided HFR video reconstruction task, we choose TimeLens (\texttt{CVPR21}\cite{Timelens}) and E-VFIA (\texttt{ICRA23}\cite{E-VFIA}). 
(3) For the event-aided image deblurring task, we choose EDI (\texttt{CVPR19}\cite{edi-cvpr19}, \texttt{TPAMI20}\cite{edi-tpami20}), LEDVDI (\texttt{ECCV20}\cite{lin2020ledvdi}), RED-Net (\texttt{ICCV21}\cite{RED-Net}), D2Net (\texttt{ICCV21}\cite{D2Net}), EVDI (\texttt{CVPR22}\cite{EVDI}), EFNet (\texttt{ECCV22}\cite{EFNet}), and NEST (\texttt{ECCV22}\cite{Teng2022NEST}). 
(4) For the event-aided image SR reconstruction task, we choose E2SRI (\texttt{CVPR20}\cite{e2sri-cvpr20}, \texttt{TPAMI22}\cite{E2SRI_pami}) and EvIntSR (\texttt{ICCV21}\cite{Han-iccv21}). 
(5) For the event-aided HDR image reconstruction task, we choose HDRev (\texttt{CVPR23}\cite{Yang_2023_HDR}) and NeuImg-HDR (\texttt{CVPR20}\cite{han2020hdr}, \texttt{TPAMI23}\cite{Han_HDR_pami}). For algorithms with both conference and journal papers, we chose their latest version for evaluation. We use the original code and pre-trained model of each method released from their project websites.

For the first four tasks (\ie{}, event-based video reconstruction, event-aided HFR reconstruction, image deblurring, and image SR), because we collect the real-captured ground truth, the results can be evaluated with the full reference evaluation metrics. We adopt PSNR to approximate estimate the human perception of reconstruction quality, and the perceptually-motivated error metric SSIM to evaluate the similarity of two images from the luminance, contrast, and structure components. We further use LPIPS \cite{LPIPS}, which better models the humane judgment by extracting the features from the pre-trained classification network, to evaluate the perceptual similarity between results and the ground truth.

\subsection{Benchmarking for event-aided methods}

We show the quantitative results on both real and simulated data for event-based video reconstruction, event-aided HFR video reconstruction, image deblurring, and image SR tasks in \fref{fig:radar} through radar plots. Representative qualitative results are presented in \fref{fig:recons}, \fref{fig:hfr}, \fref{fig:deblur}, and \fref{fig:sr} respectively. More image and video results are included in the document and video files of the supplementary material\footnote{In the supplementary material, Sec. \textcolor{red}{1} to Sec. \textcolor{red}{5} of the document file respectively shows more comparison results for five tasks on the real-captured \textsc{EventAid} dataset and the simulated \textsc{EventAid-V2E/-VM} datasets, Sec. \textcolor{red}{6} shows the distribution of quantitative results across all frames by boxplots, and the video results from different comparison methods are provided in the video file.}.

\begin{figure*}[t]
    \centering
    \includegraphics[width=1\linewidth]{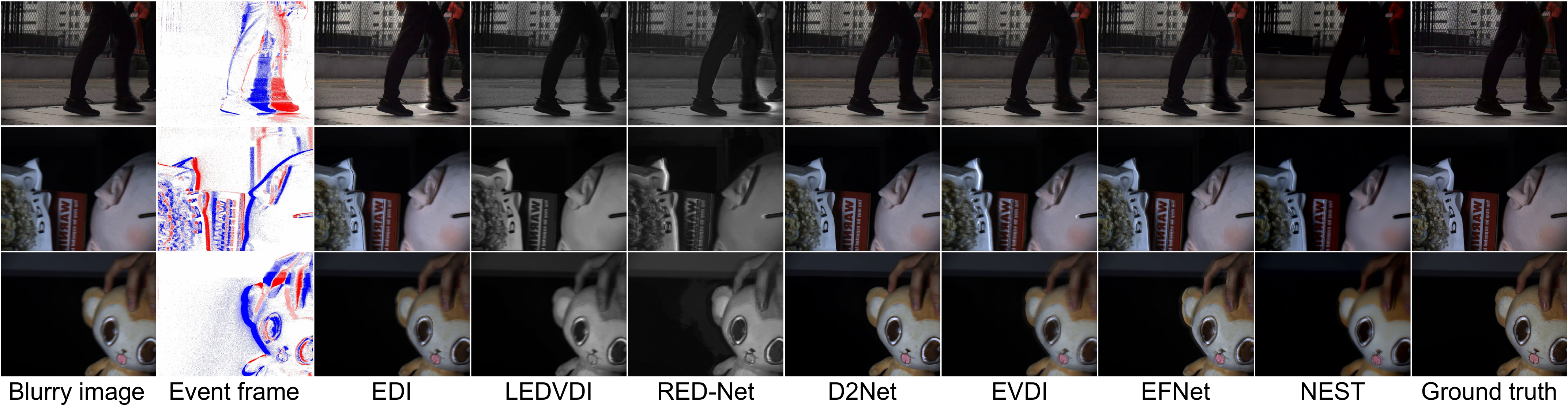}
    \caption{Event-aided image deblurring result examples from the \textsc{EventAid-B} dataset. We show the results produced from EDI \cite{edi-cvpr19}, LEDVDI \cite{lin2020ledvdi}, RED-Net \cite{RED-Net}, D2Net \cite{D2Net}, EVDI \cite{EVDI}, EFNet \cite{EFNet}, and NEST \cite{Teng2022NEST}.}
    \label{fig:deblur}
\end{figure*}

\subsubsection{Event-based video reconstruction}
We feed the input events of \textsc{EventAid-R} into the methods to be benchmarked and each of them reconstructs a video with a frame rate of $150FPS$ with timestamps matching the ground truth video. The quantitative comparison result in \fref{fig:radar} shows that ET-Net \cite{etnet} achieves more promising performance than other methods, the quantitative result distribution\footnote{The quantitative result distribution is shown in Fig. \textcolor{red}{S6-1} of the supplementary material.} also shows that ET-Net \cite{etnet} performs optimally in most groups. Nevertheless, the qualitative results of E2VID \cite{e2vid-cvpr19, e2vid-pami19} and FireNet \cite{FireNet} in \fref{fig:recons} seem more natural, retaining more detail, while other methods tend to reconstruct images with sharper edges and high contrast. It is consistent with the results in their original papers. 

\customparagraph{Inspiration:} There are two main goals in the current research for event-based video reconstruction, one is to faithfully restore details and contrasts of natural images (\eg{}, E2VID \cite{e2vid-pami19} and FireNet \cite{FireNet}), and the other one is to enhance the contrast and sharpness as well as suppress noises of images (\eg{}, ET-Net \cite{etnet} and SSL-E2VID \cite{ssl-e2vid}) to highlight the main objects of the scene. Current metrics for evaluating reconstruction quality such as PSNR only focus on global pixel value similarity, and ignore evaluating the image contrast enhancement and noise suppression quality. To comprehensively evaluate the reconstruction performance of different methods, developing new metrics for balancing detail recovery and noise suppression in event-based video reconstruction is necessary for future research. In addition, current research focuses on video reconstruction of event data from DAVIS240 \cite{davis240} or DAVIS346 \cite{davis346}, and event cameras with higher resolution (\eg{}, Prophesee EVK4 HD) also become popular. The comparison results show that some methods perform much better on the DAVIS-captured event data in the original papers ((\eg{}, Fig. 6 in EVSNN \cite{EVSNN})) than on the Prophesee-captured event data (\ie{}, \textsc{EventAid-R}) in this paper. How to reconstruct high-quality video given event data with more pixels or introduce image pre-training models to improve the reconstruction performance is worth further exploration.

\subsubsection{Event-aided HFR video reconstruction}
We extract frames from the HFR ground truth videos in \textsc{EventAid-F} by a factor of 1/8 as the LFR input videos, then feed the input events and videos into the methods to be benchmarked to reconstruct $8\times$ HFR videos. Note that in this task we only consider inter-frame interpolation and not intra-frame interpolation. We classify intra-frame interpolation into the image deblurring task. The quantitative comparison results in \fref{fig:radar}, the qualitative results shown in \fref{fig:hfr}, and additional results\footnote{More results are shown in Sec. \textcolor{red}{2.1} and Sec. \textcolor{red}{6.4} of the supplementary material.} all show the best frame interpolation performance of TimeLens \cite{Timelens}, validating its better general performance than the other method. However, there are still obvious artifacts in some qualitative results, such as distorted edges and inaccurate color recovery. 

\customparagraph{Inspiration:} Introducing events into this task mainly aims to use the high-temporal precision motion information recorded by events to restore the motion trajectory of objects, and events are mostly desired when there are non-linear and complex motions in the scene. However, existing algorithms do not always extract the motion information precisely, resulting in distorted edges and inaccurate color recovery in the interpolated frames (\eg{}, the restored colorchecker and umbrella in \fref{fig:hfr}). Besides, event degradations such as noise, tailing, and signal loss also affect the accuracy of motion extraction and the quality of image detail recovery. How to model and represent non-linear motion while eliminating the interference caused by event degradations is the main bottleneck encountered in this task. With future progress in the accurate motion extraction of event data, the effect of HFR reconstruction is expected to be further improved.

\subsubsection{Event-aided image deblurring}
We feed the input blur image sequence and corresponding events within its exposure period into deblurring methods to restore clear images at the input image's exposure period. Since LEDVDI \cite{lin2020ledvdi} and RED-Net \cite{RED-Net} can only process grayscale images, we evaluate the output results with ground truth in grayscale space. The PSNR and SSIM comparison result\footnote{The result distribution plots are shown in Sec. \textcolor{red}{6.7} of the supplementary material.} in \fref{fig:radar} shows that the performance of LEDVDI \cite{lin2020ledvdi}, EVDI \cite{EVDI}, EFNet \cite{EFNet} are all competitive, which benefit from their well-designed network model such as self-supervised strategy \cite{EVDI} and cross-modal attention \cite{EFNet}. However, the LPIPS comparison and the qualitative comparisons in \fref{fig:deblur} signally show that D2Net \cite{D2Net} can reconstruct clear and sharp images and avoid artifacts better. Compared with other algorithms, D2Net \cite{D2Net} uses the bidirectional LSTM \cite{phased-lstm-nips16} model to cleverly take the clear frame in neighborhood frames as the prior information, thus obtaining satisfactory deblurring performance. In contrast, the visual results of other methods suffer from afterimages, blurry edges, or color anomalies.

\begin{table}[t]
  \centering
  \renewcommand\arraystretch{1.7}
  \caption{The image deblurring quantitative results on the dataset collected from the controlled experiment (\fref{fig:control}). We calculate the PSNR and SSIM for 7 event-aided image deblurring methods that perform the deblurring process on input images with 11 blur levels. The blur levels correspond to the swing periods, a shorter period indicates a higher degree of blur. ``Ave.'' indicates the average values, redder blocks represent better performance with a higher PSNR value or lower LPIPS value. The values in yellow/green blocks represent the differences from average values, greener blocks represent better performance.}
  \resizebox{\hsize}{!}{
  \Huge
    \begin{tabular}{l|c|ccccccccccc}
    \toprule
    \multicolumn{13}{c}{\textbf{PSNR}} \\
    \midrule
    \multicolumn{1}{c|}{\textbf{Methods}} & \textbf{Ave.} & \multicolumn{1}{c}{1.50s} & \multicolumn{1}{c}{1.75s} & \multicolumn{1}{c}{2.00s} & \multicolumn{1}{c}{2.25s} & \multicolumn{1}{c}{2.50s} & \multicolumn{1}{c}{2.75s} & \multicolumn{1}{c}{3.00s} & \multicolumn{1}{c}{3.25s} & \multicolumn{1}{c}{3.50s} & \multicolumn{1}{c}{3.75s} & \multicolumn{1}{c}{4.00s} \\
    \midrule
    EDI\cite{edi-tpami20}   & \cellcolor[rgb]{ .988,  .988,  1} 16.67  & \cellcolor[rgb]{ .388,  .745,  .482} +0.40 & \cellcolor[rgb]{ .667,  .835,  .541} +0.14 & \cellcolor[rgb]{ .698,  .843,  .549} +0.11 & \cellcolor[rgb]{ .882,  .902,  .588} -0.07 & \cellcolor[rgb]{ .804,  .878,  .573} +0.01 & \cellcolor[rgb]{ 1,  .937,  .612} -0.18 & \cellcolor[rgb]{ .859,  .894,  .584} -0.05 & \cellcolor[rgb]{ .906,  .91,  .592} -0.09 & \cellcolor[rgb]{ .875,  .898,  .588} -0.06 & \cellcolor[rgb]{ .839,  .89,  .58} -0.03 & \cellcolor[rgb]{ .996,  .937,  .612} -0.18 \\
    LEDVDI\cite{lin2020ledvdi} & \cellcolor[rgb]{ .973,  .412,  .42} 32.02  & \cellcolor[rgb]{ 1,  .937,  .612} -0.33 & \cellcolor[rgb]{ .78,  .871,  .569} -0.15 & \cellcolor[rgb]{ .616,  .82,  .533} -0.01 & \cellcolor[rgb]{ .694,  .843,  .549} -0.07 & \cellcolor[rgb]{ .388,  .745,  .482} +0.18 & \cellcolor[rgb]{ .459,  .769,  .498} +0.13 & \cellcolor[rgb]{ .631,  .824,  .537} -0.02 & \cellcolor[rgb]{ .576,  .804,  .525} +0.03 & \cellcolor[rgb]{ .537,  .792,  .514} +0.06 & \cellcolor[rgb]{ .573,  .804,  .522} +0.03 & \cellcolor[rgb]{ .427,  .761,  .494} +0.15 \\
    RED-Net\cite{RED-Net} & \cellcolor[rgb]{ .984,  .804,  .816} 21.62  & \cellcolor[rgb]{ 1,  .937,  .612} -2.21 & \cellcolor[rgb]{ .871,  .898,  .584} -1.37 & \cellcolor[rgb]{ .647,  .827,  .537} +0.08 & \cellcolor[rgb]{ .6,  .812,  .529} +0.37 & \cellcolor[rgb]{ .49,  .776,  .506} +1.08 & \cellcolor[rgb]{ .388,  .745,  .482} +1.71 & \cellcolor[rgb]{ .604,  .816,  .529} +0.34 & \cellcolor[rgb]{ .62,  .82,  .533} +0.25 & \cellcolor[rgb]{ .616,  .82,  .533} +0.26 & \cellcolor[rgb]{ .6,  .812,  .529} +0.36 & \cellcolor[rgb]{ .792,  .875,  .569} -0.86 \\
    D2Net\cite{D2Net} & \cellcolor[rgb]{ .976,  .533,  .545} 28.80  & \cellcolor[rgb]{ .388,  .745,  .482} +0.91 & \cellcolor[rgb]{ .549,  .796,  .518} +0.56 & \cellcolor[rgb]{ .686,  .839,  .545} +0.27 & \cellcolor[rgb]{ .757,  .863,  .561} +0.11 & \cellcolor[rgb]{ .82,  .882,  .576} -0.02 & \cellcolor[rgb]{ .894,  .906,  .592} -0.18 & \cellcolor[rgb]{ .941,  .918,  .6} -0.28 & \cellcolor[rgb]{ .929,  .918,  .6} -0.26 & \cellcolor[rgb]{ .961,  .925,  .604} -0.32 & \cellcolor[rgb]{ 1,  .937,  .612} -0.41 & \cellcolor[rgb]{ .988,  .937,  .612} -0.39 \\
    EVDI\cite{EVDI}  & \cellcolor[rgb]{ .976,  .451,  .459} 31.05  & \cellcolor[rgb]{ 1,  .937,  .612} -0.29 & \cellcolor[rgb]{ .875,  .898,  .588} -0.16 & \cellcolor[rgb]{ .878,  .902,  .588} -0.16 & \cellcolor[rgb]{ .941,  .922,  .6} -0.23 & \cellcolor[rgb]{ .729,  .855,  .557} -0.01 & \cellcolor[rgb]{ .792,  .875,  .569} -0.07 & \cellcolor[rgb]{ .608,  .816,  .529} +0.12 & \cellcolor[rgb]{ .541,  .792,  .518} +0.19 & \cellcolor[rgb]{ .541,  .796,  .518} +0.19 & \cellcolor[rgb]{ .655,  .831,  .541} +0.07 & \cellcolor[rgb]{ .388,  .745,  .482} +0.35 \\
    EFNet\cite{EFNet} & \cellcolor[rgb]{ .976,  .51,  .522} 29.41  & \cellcolor[rgb]{ 1,  .937,  .612} -0.70 & \cellcolor[rgb]{ .929,  .918,  .6} -0.55 & \cellcolor[rgb]{ .894,  .906,  .592} -0.48 & \cellcolor[rgb]{ .847,  .89,  .58} -0.38 & \cellcolor[rgb]{ .722,  .851,  .553} -0.12 & \cellcolor[rgb]{ .714,  .847,  .553} -0.10 & \cellcolor[rgb]{ .498,  .78,  .506} +0.36 & \cellcolor[rgb]{ .412,  .753,  .49} +0.54 & \cellcolor[rgb]{ .424,  .757,  .49} +0.51 & \cellcolor[rgb]{ .506,  .784,  .51} +0.34 & \cellcolor[rgb]{ .388,  .745,  .482} +0.58 \\
    NEST\cite{Teng2022NEST}  & \cellcolor[rgb]{ .988,  .871,  .882} 19.84  & \cellcolor[rgb]{ 1,  .937,  .612} -0.04 & \cellcolor[rgb]{ .745,  .859,  .561} +0.01 & \cellcolor[rgb]{ .804,  .878,  .573} -0.00 & \cellcolor[rgb]{ .831,  .886,  .576} -0.01 & \cellcolor[rgb]{ .388,  .745,  .482} +0.08 & \cellcolor[rgb]{ .824,  .882,  .576} -0.01 & \cellcolor[rgb]{ .945,  .922,  .6} -0.03 & \cellcolor[rgb]{ .816,  .882,  .573} -0.01 & \cellcolor[rgb]{ .678,  .835,  .545} +0.02 & \cellcolor[rgb]{ .643,  .827,  .537} +0.03 & \cellcolor[rgb]{ .925,  .914,  .596} -0.03 \\
    \midrule
    \multicolumn{13}{c}{\textbf{LPIPS}} \\
    \midrule
    \multicolumn{1}{c|}{\textbf{Methods}} & \textbf{Ave.} & \multicolumn{1}{c}{1.50s} & \multicolumn{1}{c}{1.75s} & \multicolumn{1}{c}{2.00s} & \multicolumn{1}{c}{2.25s} & \multicolumn{1}{c}{2.50s} & \multicolumn{1}{c}{2.75s} & \multicolumn{1}{c}{3.00s} & \multicolumn{1}{c}{3.25s} & \multicolumn{1}{c}{3.50s} & \multicolumn{1}{c}{3.75s} & \multicolumn{1}{c}{4.00s} \\
    \midrule
    EDI\cite{edi-tpami20}   & \cellcolor[rgb]{ .988,  .988,  1} 0.395  & \cellcolor[rgb]{ 1,  .937,  .612} +0.034 & \cellcolor[rgb]{ .961,  .925,  .604} +0.030 & \cellcolor[rgb]{ .882,  .898,  .584} +0.022 & \cellcolor[rgb]{ .816,  .878,  .573} +0.015 & \cellcolor[rgb]{ .659,  .827,  .537} -0.001 & \cellcolor[rgb]{ .71,  .843,  .549} +0.004 & \cellcolor[rgb]{ .557,  .796,  .518} -0.012 & \cellcolor[rgb]{ .439,  .761,  .49} -0.024 & \cellcolor[rgb]{ .455,  .765,  .494} -0.023 & \cellcolor[rgb]{ .506,  .78,  .506} -0.017 & \cellcolor[rgb]{ .388,  .745,  .482} -0.030 \\
    LEDVDI\cite{lin2020ledvdi} & \cellcolor[rgb]{ .973,  .427,  .435} 0.258  & \cellcolor[rgb]{ .792,  .871,  .565} +0.003 & \cellcolor[rgb]{ .957,  .922,  .6} +0.013 & \cellcolor[rgb]{ .922,  .91,  .592} +0.011 & \cellcolor[rgb]{ 1,  .937,  .612} +0.015 & \cellcolor[rgb]{ .812,  .878,  .573} +0.004 & \cellcolor[rgb]{ .757,  .859,  .557} +0.000 & \cellcolor[rgb]{ .647,  .824,  .537} -0.007 & \cellcolor[rgb]{ .639,  .824,  .533} -0.007 & \cellcolor[rgb]{ .667,  .831,  .541} -0.005 & \cellcolor[rgb]{ .698,  .839,  .545} -0.003 & \cellcolor[rgb]{ .388,  .745,  .482} -0.023 \\
    RED-Net\cite{RED-Net} & \cellcolor[rgb]{ .984,  .882,  .894} 0.370  & \cellcolor[rgb]{ 1,  .937,  .612} +0.051 & \cellcolor[rgb]{ .941,  .918,  .596} +0.043 & \cellcolor[rgb]{ .792,  .871,  .565} +0.023 & \cellcolor[rgb]{ .757,  .859,  .561} +0.018 & \cellcolor[rgb]{ .596,  .808,  .525} -0.005 & \cellcolor[rgb]{ .537,  .792,  .514} -0.013 & \cellcolor[rgb]{ .529,  .788,  .51} -0.014 & \cellcolor[rgb]{ .439,  .761,  .49} -0.026 & \cellcolor[rgb]{ .439,  .761,  .49} -0.026 & \cellcolor[rgb]{ .49,  .776,  .502} -0.019 & \cellcolor[rgb]{ .388,  .745,  .482} -0.033 \\
    D2Net\cite{D2Net} & \cellcolor[rgb]{ .973,  .412,  .42} 0.254  & \cellcolor[rgb]{ .388,  .745,  .482} -0.031 & \cellcolor[rgb]{ .706,  .843,  .549} -0.007 & \cellcolor[rgb]{ .753,  .859,  .557} -0.004 & \cellcolor[rgb]{ .82,  .878,  .573} +0.001 & \cellcolor[rgb]{ .875,  .898,  .584} +0.005 & \cellcolor[rgb]{ .784,  .867,  .565} -0.002 & \cellcolor[rgb]{ .843,  .886,  .576} +0.002 & \cellcolor[rgb]{ .992,  .933,  .608} +0.013 & \cellcolor[rgb]{ 1,  .937,  .612} +0.014 & \cellcolor[rgb]{ .984,  .929,  .608} +0.013 & \cellcolor[rgb]{ .749,  .859,  .557} -0.004 \\
    EVDI\cite{EVDI}  & \cellcolor[rgb]{ .976,  .569,  .576} 0.293  & \cellcolor[rgb]{ 1,  .937,  .612} +0.032 & \cellcolor[rgb]{ .98,  .929,  .604} +0.030 & \cellcolor[rgb]{ .867,  .894,  .58} +0.018 & \cellcolor[rgb]{ .863,  .894,  .58} +0.018 & \cellcolor[rgb]{ .69,  .839,  .545} -0.000 & \cellcolor[rgb]{ .659,  .827,  .537} -0.004 & \cellcolor[rgb]{ .573,  .8,  .522} -0.013 & \cellcolor[rgb]{ .486,  .773,  .502} -0.022 & \cellcolor[rgb]{ .522,  .784,  .51} -0.018 & \cellcolor[rgb]{ .608,  .812,  .525} -0.009 & \cellcolor[rgb]{ .388,  .745,  .482} -0.032 \\
    EFNet\cite{EFNet} & \cellcolor[rgb]{ .973,  .475,  .482} 0.270  & \cellcolor[rgb]{ .745,  .855,  .557} +0.000 & \cellcolor[rgb]{ .965,  .925,  .604} +0.011 & \cellcolor[rgb]{ .89,  .902,  .588} +0.008 & \cellcolor[rgb]{ 1,  .937,  .612} +0.013 & \cellcolor[rgb]{ .894,  .902,  .588} +0.008 & \cellcolor[rgb]{ .765,  .863,  .561} +0.001 & \cellcolor[rgb]{ .545,  .792,  .514} -0.010 & \cellcolor[rgb]{ .51,  .78,  .506} -0.011 & \cellcolor[rgb]{ .635,  .82,  .533} -0.005 & \cellcolor[rgb]{ .769,  .863,  .561} +0.002 & \cellcolor[rgb]{ .388,  .745,  .482} -0.018 \\
    NEST\cite{Teng2022NEST}  & \cellcolor[rgb]{ .984,  .933,  .945} 0.382  & \cellcolor[rgb]{ .486,  .776,  .502} -0.005 & \cellcolor[rgb]{ .459,  .765,  .494} -0.005 & \cellcolor[rgb]{ .4,  .745,  .482} -0.007 & \cellcolor[rgb]{ .757,  .859,  .557} +0.004 & \cellcolor[rgb]{ .541,  .792,  .514} -0.003 & \cellcolor[rgb]{ .388,  .745,  .482} -0.008 & \cellcolor[rgb]{ .58,  .804,  .522} -0.002 & \cellcolor[rgb]{ .925,  .914,  .596} +0.009 & \cellcolor[rgb]{ .922,  .91,  .592} +0.009 & \cellcolor[rgb]{ 1,  .937,  .612} +0.011 & \cellcolor[rgb]{ .545,  .792,  .514} -0.003 \\
    \bottomrule
    \end{tabular}}%
  \label{tab:control}%
  \vspace{-10pt}
\end{table}%

\customparagraph{controlled experiment:} We further conduct an experiment for Event-aided image deblurring to evaluate the limits of blur levels that existing algorithms can withstand. \Tref{tab:control} records the quantitative results\footnote{The visual deblurring results of different methods are shown in Fig. \textcolor{red}{S3-31} and Fig. \textcolor{red}{S3-32} of the supplementary material.}. Consistent with the above conclusions, the average PSNR values of LEDVDI \cite{lin2020ledvdi}, EVDI \cite{EVDI}, EFNet \cite{EFNet} are more prominent, while D2Net \cite{D2Net} is significantly superior to other algorithms in LPIPS and visual performance. The changing trend of the colored blocks in \Tref{tab:control} reveals the algorithm's tolerance and compatibility with different degrees of blur. The performance of most algorithms improves as the degree of blur decreases, but there are also a few algorithms that have the opposite trend. This counterintuitive result may be caused by the relatively uniform degree of blur in the training dataset for learning-based algorithms. As a non-learning-based method, when EDI \cite{edi-cvpr19, edi-tpami20} is dealing with a lower degree of blur, the noise introduced by event data reduces the quality of the reconstructed image.

\customparagraph{Inspiration:} From the comparison in \fref{fig:deblur} and \Tref{tab:control}, we can find that there are two issues that prevent existing methods from being further improved: 
The modal differences between events and images and the difficulty in calibrating event trigger thresholds lead to additional artifacts introduced by event signals. Designing an event representation model that is more effective for image fusion, and proposing a robust online threshold estimation method might help in conquering these bottlenecks.
Besides, similar to the HFR reconstruction task, the difficulty of extracting the non-linear motion accurately makes it hard to restore sharp textures correctly. So precise motion extractions are also desired here. 
In addition, event cameras perceive in grayscale space, so it is difficult to assist in recovering the color information of blurry areas. Introducing color event cameras or image colorization models are possible ways of improvement.
Finally, due to the large spatial resolution gap between event cameras and frame cameras, how to use LR event signals to deblur HR images with much higher resolution (like 20 times larger) is of practical value for enhancing the photography experience in future camera phones.

\begin{figure}[t]
    \centering
    \includegraphics[width=1\linewidth]{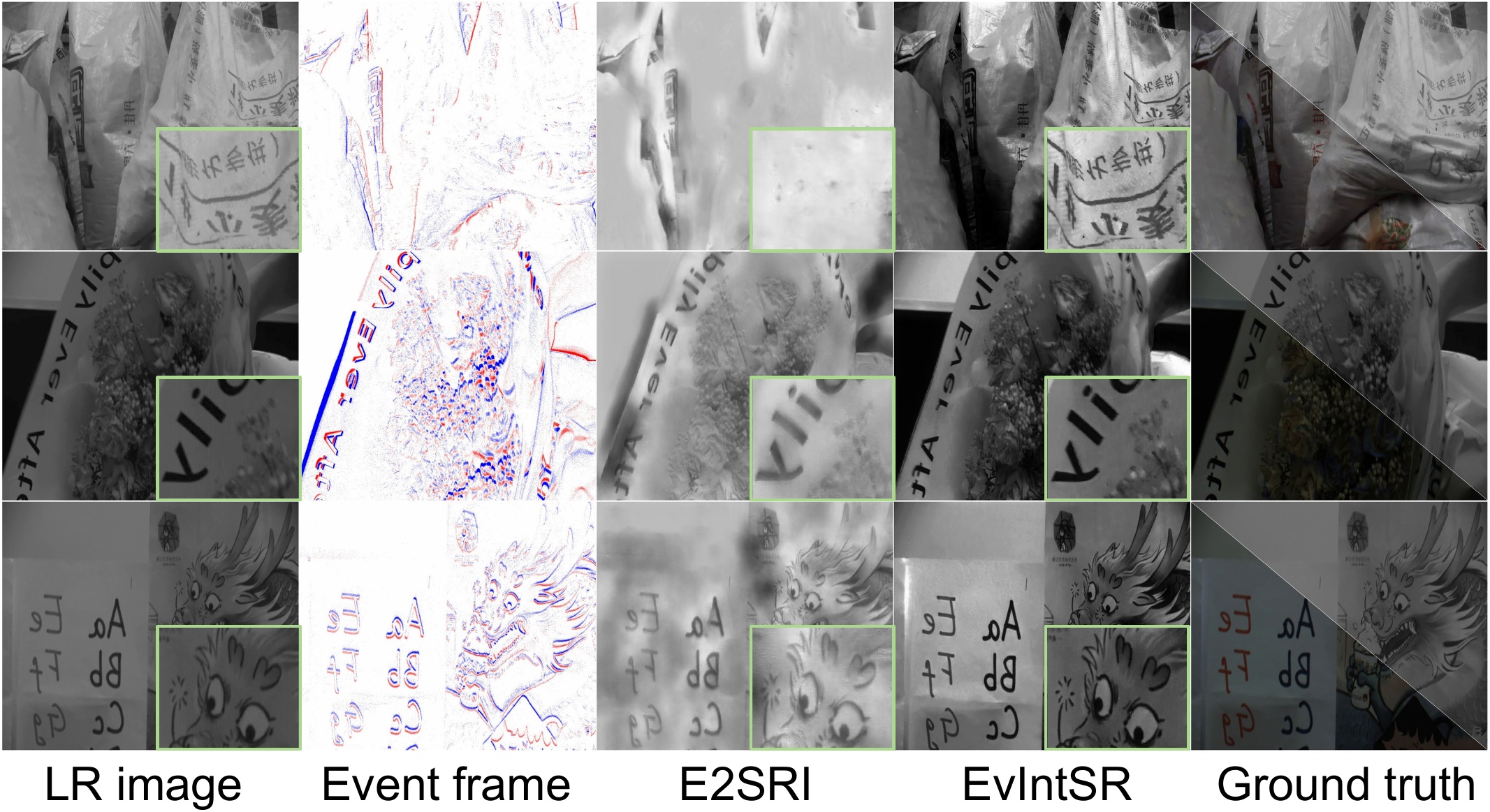}
    \caption{Event-aided image super-resolution result examples from the \textsc{EventAid-S} dataset. We show the results produced from E2SRI \cite{e2sri-cvpr20, E2SRI_pami}, and EvIntSR \cite{Han-iccv21}. Green boxes show the closed-up views.}
    \label{fig:sr}
    \vspace{-10pt}
\end{figure}

\begin{figure*}[t]
    \centering
    \includegraphics[width=1\linewidth]{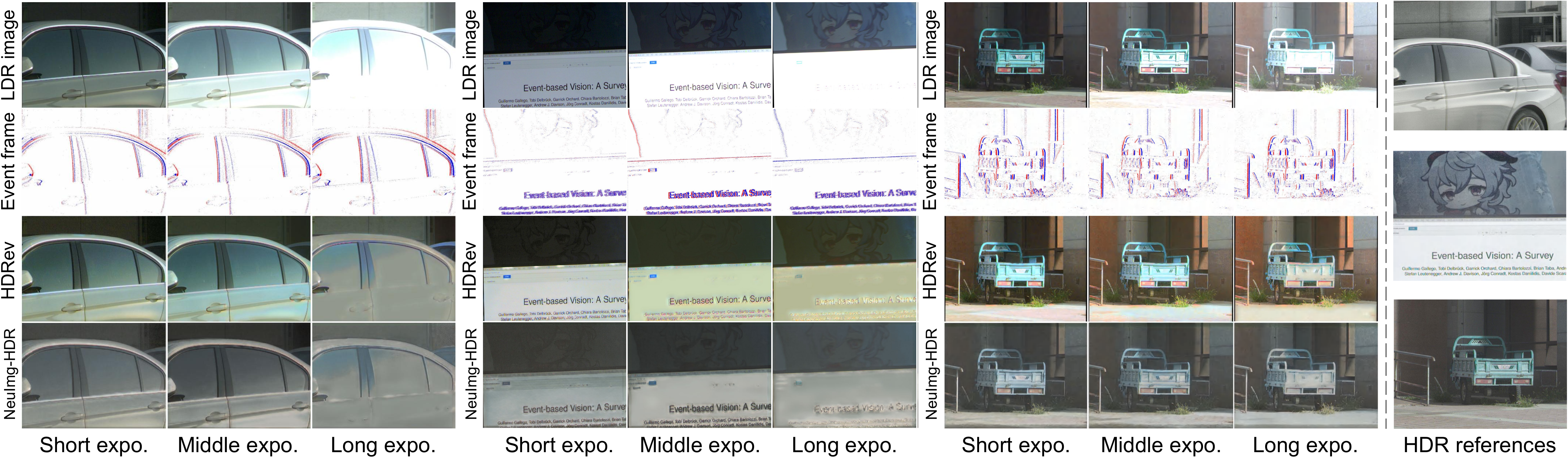}
    \caption{Event-aided HDR image reconstruction result examples from the \textsc{EventAid-D} dataset. We show the results reconstructed from HDRev \cite{Yang_2023_HDR} and NeuImg-HDR \cite{han2020hdr, Han_HDR_pami} for short-/middle-/long-exposure LDR images. The rightmost column shows the HDR reference, which is merged from 20 multi-exposure LDR images.}
    \label{fig:hdr}
    \vspace{-10pt}
\end{figure*}

\subsubsection{Event-aided image super-resolution}
We generate input $1\times$ LR images by downsampling HR images of \textsc{EventAid-S} at a factor of 1/2 following the process in single image SR tasks, then feed the $1\times$ LR image sequences and events into selected methods to reconstruct $2\times$ SR frames. Note that E2SRI \cite{e2sri-cvpr20, e2vid-pami19} executes the SR process directly from pure event data, so we only feed $1\times$ LR events into it. We evaluate the output results with ground truth in grayscale space since E2SRI \cite{e2sri-cvpr20, e2vid-pami19} and EvIntSR \cite{Han-iccv21} can only process grayscale images. EventZoom \cite{EventZoom} and ESR \cite{Weng_2022_ECCV} are not compared here because these algorithms output high-resolution event signals rather than directly output high-resolution images. The quantitative and qualitative comparison results\footnote{More results are shown in Sec. \textcolor{red}{4.1} and Sec. \textcolor{red}{6.10} of the supplementary material.} in \fref{fig:radar} and \fref{fig:sr} show the best frame SR performance of EvIntSR \cite{Han-iccv21}. However, the comparison can not validate that the performance of EvIntSR \cite{Han-iccv21} is better than the other method because the input data of the two methods are inconsistent. For EvIntSR \cite{Han-iccv21} with only input events, the spatial resolution cannot be amplified by events when the scene is stationary because no events are triggered.

\customparagraph{Inspiration:} The core principle of this task is to use the high-temporal precision motion information recorded by events to convert sub-pixel displacements in the spatial domain, thereby achieving spatial upsampling of images. Pure image-based SR has been studied for decades, and some learning-based methods can even achieve $16\times$ upsampling (\eg{}, ABPN \cite{Liu2019abpn}), while event-aided methods can only achieve lower-factor upsampling, such as $2\times/4\times$ SR. The rationale for introducing events into the image SR task is that the prior information provided by events has the potential (compared to pure learning-based methods in the image domain) to reconstruct reliable sub-pixel information. New explorations should make better use of events to reconstruct SR images of higher quality than pure image-based methods to further demonstrate the practical value and research significance of this event-aided task.

\subsubsection{Event-aided HDR image reconstruction}
We feed the input LDR images and corresponding events within into the selected HDRev \cite{Yang_2023_HDR} and NeuImg-HDR \cite{han2020hdr, Han_HDR_pami} to restore HDR images. We obtain LDR images captured through short-/middle-/long-exposure by alternating exposure and used them as input to fully test the robustness of methods to two different types of LDR images, \ie{}, under-/over-exposure. To obtain HDR reference as much as possible, we first hold the scene still while shooting the data and capture 11 multi-exposure frames to synthesize the reference image by Debevec \etal \cite{mutliHDR}. \Fref{fig:hdr} shows the HDR restoration result examples\footnote{More additional results are shown in Sec. \textcolor{red}{5} of the supplementary material.} of the two methods for short-/middle-/long-exposure cases. It can be seen from the comparison that both algorithms show a trend that the recovery performance of the under-exposed area is better than that of the over-exposed area, perhaps because the white background color of the over-exposed area makes it easier to highlight the reconstructed artifacts. The color of the image reconstructed by HDRev \cite{Yang_2023_HDR} is more in line with human vision, while the texture details restored by NeuImg-HDR \cite{han2020hdr, Han_HDR_pami} are clearer. 

\customparagraph{Inspiration:} Event-aided HDR image reconstruction methods use the texture motion of over-/under-exposed areas perceived by events to recover the lost information of these areas and fuse them with LDR images. Accurately restoring texture and color are two major attributes that the methods of this task need to own. The challenge of reconstructing realistic textures in over-/under-exposed areas is similar to that of the event-based video reconstruction task.
In addition, current methods for Event-aided HDR image restoration \cite{han2020hdr} require shaking the event camera when capturing data to make the event camera sufficiently perceive the texture in the scene. Breaking through this limitation will enable the algorithm to be more conveniently and broadly used. Moreover, it is difficult for existing methods to correctly restore the color of over-/under-exposed areas because neither the event nor the image provides color priors, \eg{}, the color of the over-exposed car door in the first example and the under-exposed cartoon pattern in the second example in \fref{fig:hdr} are not accurately restored. Therefore, how to recover the color of HDR areas is also a challenge that needs to be explored. Using color restoration strategies in algorithms such as image colorization, image inpainting, and semantic-based image restoration, or using existing color recovery pre-trained models might be helpful to solve the problem faced by this task.

\begin{figure*}[t]
    \centering
    \includegraphics[width=1\linewidth]{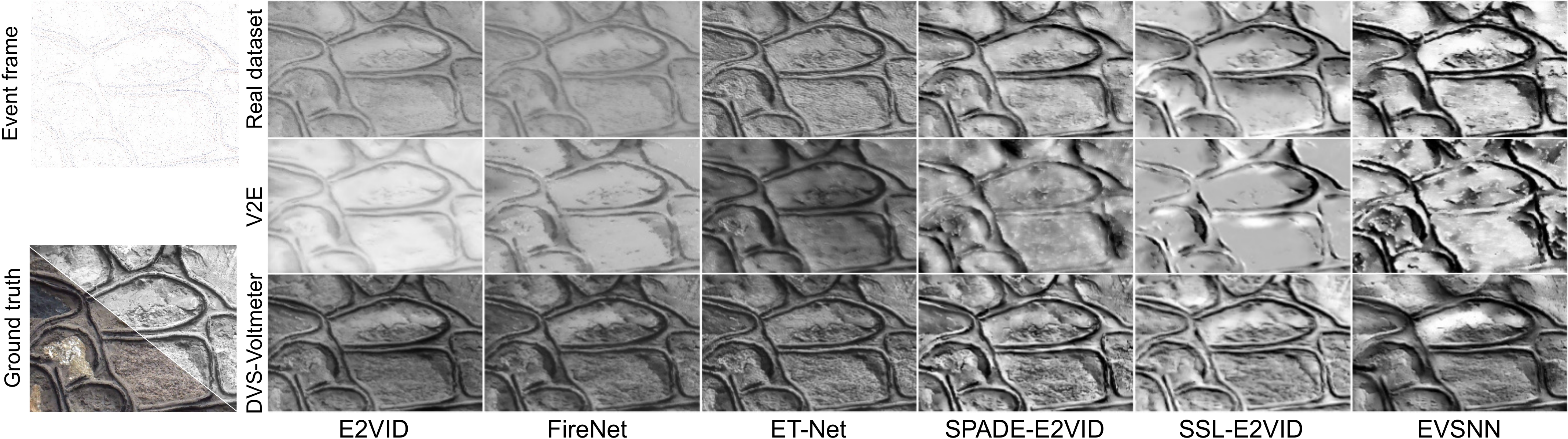}
    \caption{Event-based video reconstruction result examples from real-captured \textsc{EventAid-R}, V2E \cite{V2E} simulated dataset \textsc{EventAid-R-V2E}, and DVS-Voltmeter \cite{DVS-voltmeter} simulated dataset \textsc{EventAid-R-VM}.The reconstruction methods are E2VID \cite{e2vid-cvpr19, e2vid-pami19}, FireNet \cite{FireNet}, ET-Net \cite{etnet}, SPADE-E2VID \cite{SPADE-E2VID}, SSL-E2VID \cite{ssl-e2vid}, and EVSNN \cite{EVSNN}.}
    \label{fig:recons-sim}
\end{figure*}

\begin{figure*}[h]
    \centering
    \includegraphics[width=1\linewidth]{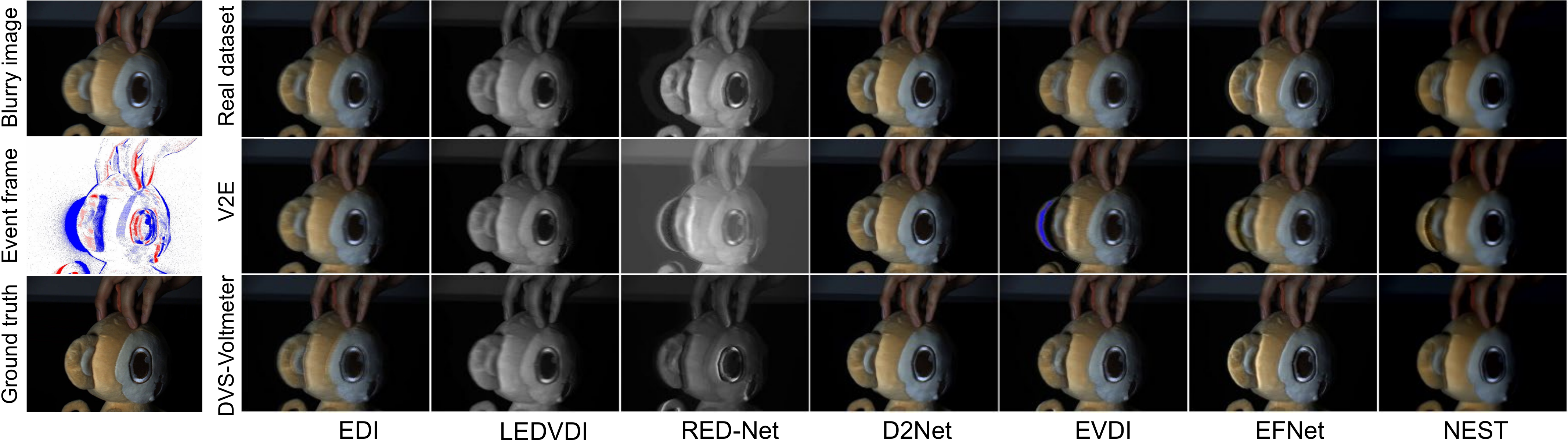}
    \caption{Event-aided image deblurring result examples from real-captured \textsc{EventAid-B}, V2E \cite{V2E} simulated dataset \textsc{EventAid-B-V2E}, and DVS-Voltmeter \cite{DVS-voltmeter} simulated dataset \textsc{EventAid-B-VM}. The deblurring methods are EDI \cite{edi-cvpr19}, LEDVDI \cite{lin2020ledvdi}, RED-Net \cite{RED-Net}, D2Net \cite{D2Net}, EVDI \cite{EVDI}, EFNet \cite{EFNet}, and NEST \cite{Teng2022NEST}.}
    \label{fig:deblur-sim}
    \vspace{-10pt}
\end{figure*}

\subsection{Comparison with event simulators}
Event simulators are useful for quickly generating large-scale training datasets, but the real-sim gap makes it difficult for trained models to work efficiently on real-captured data, which has been verified by NeuroZoom \cite{NeuroZoom}. We execute the above benchmark processing again on the simulated \textsc{EventAid-V2E/-VM} datasets to compare the construction results. The quantitative results on simulated data are recorded in \fref{fig:radar} through radar plots. Representative qualitative results on event-based video reconstruction and event-aided image deblurring tasks are presented in \fref{fig:recons-sim} and \fref{fig:deblur-sim} respectively\footnote{More frame and video results are included in the supplementary material. Sec. \textcolor{red}{6} also shows the distribution of quantitative results across all frames by boxplots.}.

The quantitative results are recorded in \fref{fig:radar}. It can be seen that the performance ranking on simulated data is close to the ranking on the real-captured dataset \textsc{EventAid}. This indirectly proves that the spatiotemporal synchronization error of our \textsc{EventAid} dataset is not significant, and the comparisons on real-captured \textsc{EventAid} are convincing. The results in \fref{fig:radar} show that the performances on V2E-simulated datasets are lower than on real data, while some methods perform against this trend, such as ET-Net \cite{etnet} in the video reconstruction task. This may be because the pre-trained models of these methods are trained on V2E-simulated datasets. In contrast, the performances on \textsc{EventAid-VM} datasets are similar or even higher than the corresponding performance on real data, especially the video reconstruction task. Because the DVS-Voltmeter \cite{DVS-voltmeter} more realistically models the triggering process of event signals, the simulated event distribution model matches the real data. In contrast, the quality of V2E \cite{V2E} depends on the frame rate of the input video. When the frame rate is low, the generated events are difficult to simulate the continuous distribution of real events in the time domain. The comparison of qualitative results also shows that \textsc{EventAid-VM} dataset results in better visual effects. It is worth noting that in the event-based video reconstruction task, the image reconstructed on the \textsc{EventAid-VM} has a greater contrast than the result on real-captured \textsc{EventAid}, which may be related to the inaccurate setting of the trigger threshold of the simulator. Whether the event simulator can accurately simulate the trigger mechanism of the real event sensor, correctly model the degradation process such as noise, trailing, and signal loss, and solve the problem of discontinuous event distribution in time dimension when converting low frame rate video into event signal will determine whether the event simulator can provide effective training and evaluation data set for the study of event algorithms.

\section{Conclusions}
We propose the first high-quality evaluation dataset for event-aided image/video enhancement tasks with real-captured data that allow quantitative and qualitative evaluations, and the first comprehensive benchmarking of the existing event-aided image/video enhancement algorithms. All data, including input events and frames and the ground truth, are real-world captured by beam splitter-mounted hybrid camera systems. We benchmark 19 state-of-the-art algorithms for both five tasks and statistically analyze their performances in different scenarios, and also benchmark 2 widely used event simulators. Finally, we discuss the performance of existing methods from the results on the \textsc{EventAid} dataset and propose several open problems for future researchers.

\customparagraph{Limitations:} Some published methods have not been benchmarked in this paper because the codes are unavailable or they have just been published. we will release a benchmark website that allows researchers to update their methods to continuously facilitate research on event-aided image/video enhancement tasks after the acceptance of this paper.

\ifCLASSOPTIONcaptionsoff
  \newpage
\fi


\begin{IEEEbiography}
[{\includegraphics[width=1in,height=1.25in,clip,keepaspectratio]{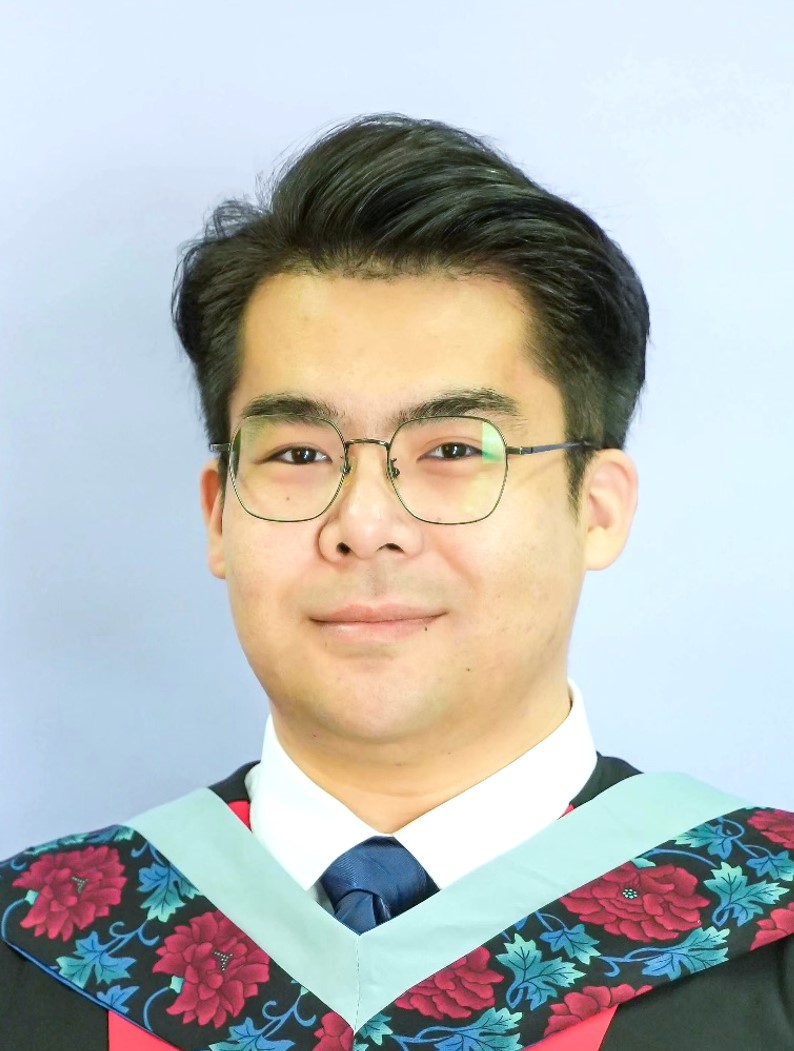}}]
{Peiqi Duan}
 is currently a Boya Postdoctoral Fellow at the School of Computer Science, Peking University. He received his PhD degree from Peking University in 2023. His research interests span event-based imaging and vision, single-image super-resolution, and HDR image reconstruction. He has served as a reviewer/program committee member for TPAMI, IJCV, TCSVT, CVPR, ICCV, ECCV, NeurIPS, etc.
\end{IEEEbiography}
\vspace{-4mm}

\begin{IEEEbiography}
[{\includegraphics[width=1in,height=1.25in,clip,keepaspectratio]{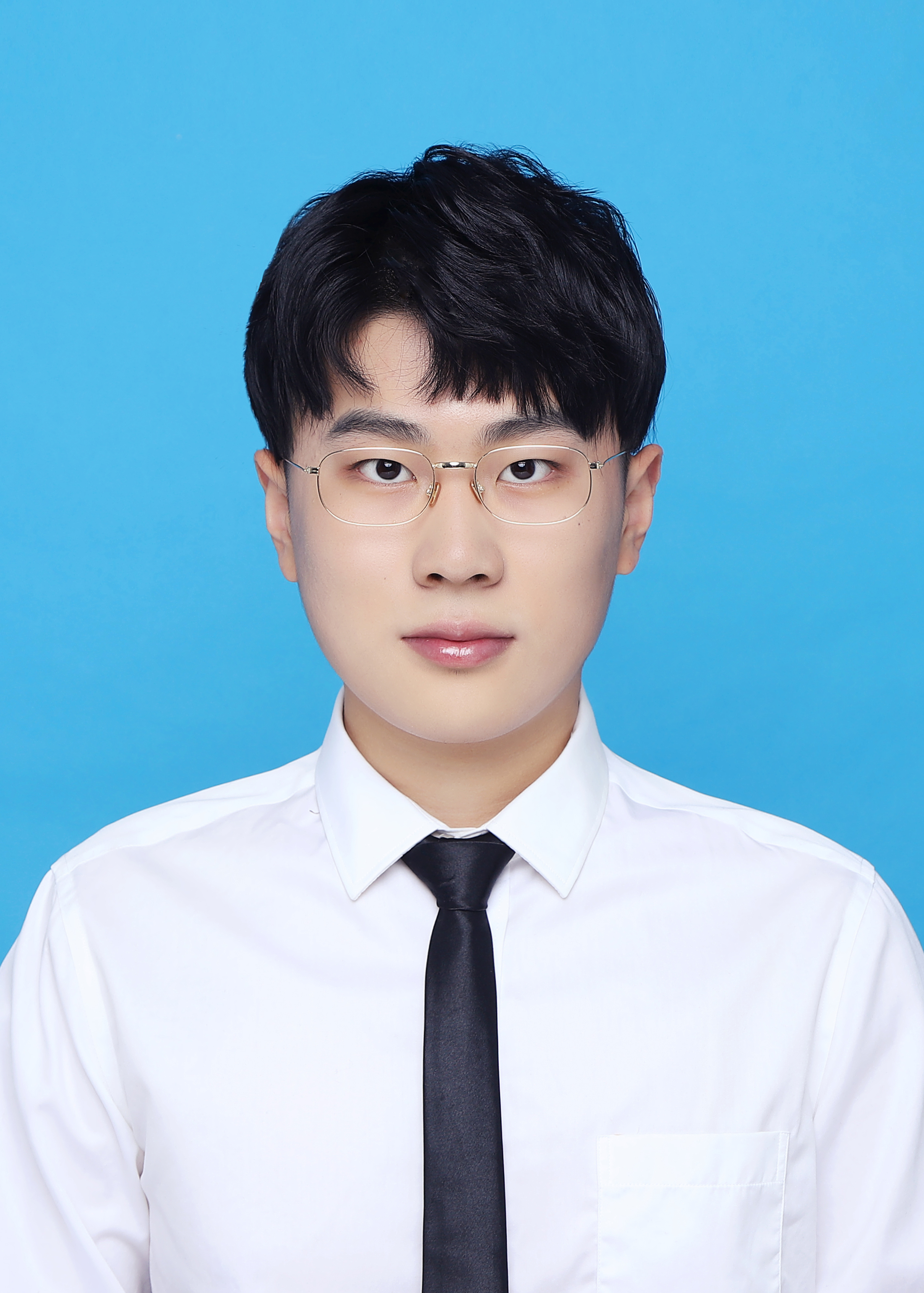}}]
{Boyu Li}
is currently a Ph.D. student in the School of Computer Science of Peking University. He received the B.S. degree from Peking University in 2023. His research interests include event-based vision and related topics.
\end{IEEEbiography}
\vspace{-4mm}

\begin{IEEEbiography}
[{\includegraphics[width=1in,height=1.25in,clip,keepaspectratio]{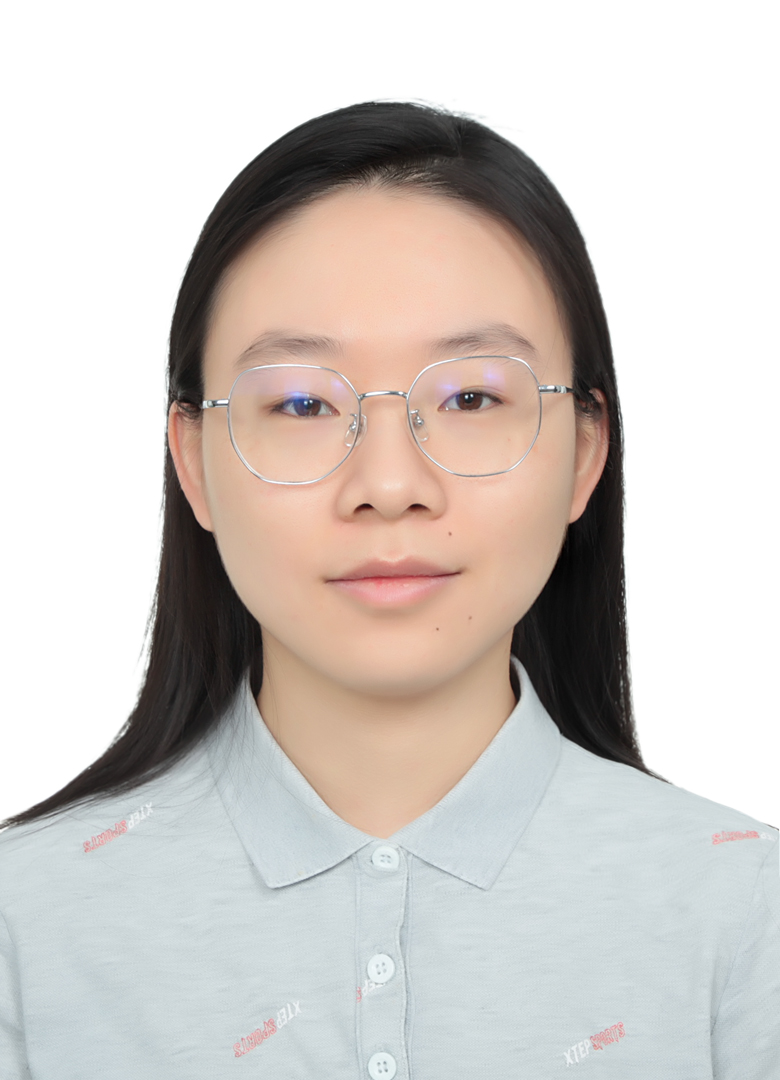}}]
{Yixin Yang}
is a Ph.D. student in the School of Computer Science, Peking University. Her research interests span event-based imaging and vision, hybrid-camera super-resolution and HDR reconstruction.

\end{IEEEbiography}
\vspace{-4mm}

\begin{IEEEbiography}
[{\includegraphics[width=1in,height=1.25in,clip,keepaspectratio]{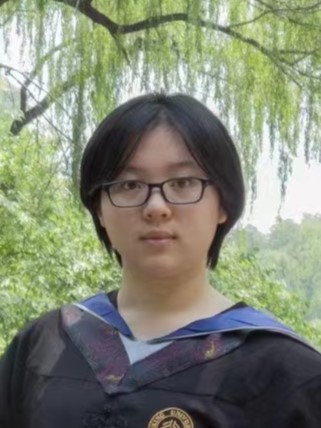}}]
{Hanyue Lou}
received the B.S. degree summa cum laude from Peking University, Beijing, China, in 2023. She is currently working toward the Ph.D. degree with the National Engineering Research Center of Video Technology, School of Computer Science, Peking University. Her research interests are focused on applications of neuromorphic cameras.
\end{IEEEbiography}
\vspace{-4mm}

\begin{IEEEbiography}[{\includegraphics[width=1in,height=1.25in,clip,keepaspectratio]{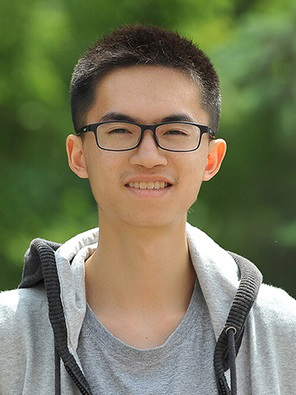}}]
{Minggui Teng}
received the B.S. degree from Peking University, Beijing, China, in 2021. He is currently working toward the Ph.D. degree with the National Engineering Research Center of Video Technology, School of Computer Science, Peking University. His research interests are focused on neuromorphic camera and image enhancement. He has served as a reviewer for CVPR, ICCV, ECCV, etc.
\end{IEEEbiography}
\vspace{-4mm}

\begin{IEEEbiography}
[{\includegraphics[width=1in,height=1.25in,clip,keepaspectratio]{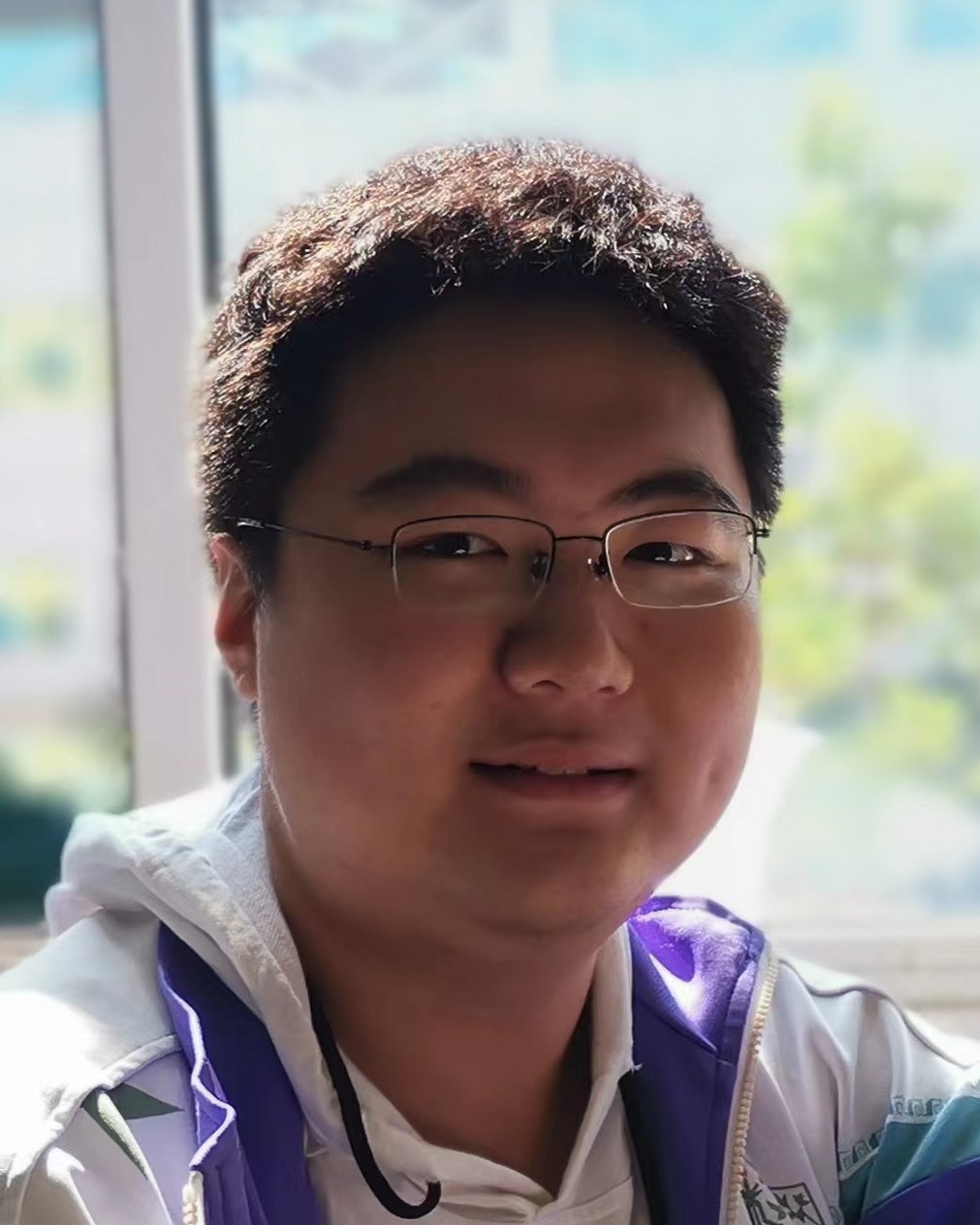}}]
{Yi Ma}
received the B.S. degree from Peking University in 2021. He is currently a graduate student at Peking University. His research interests are centered around event-based vision signal processing.
\end{IEEEbiography}
\vspace{-4mm}

\begin{IEEEbiography}
[{\includegraphics[width=1in,height=1.25in,clip,keepaspectratio]{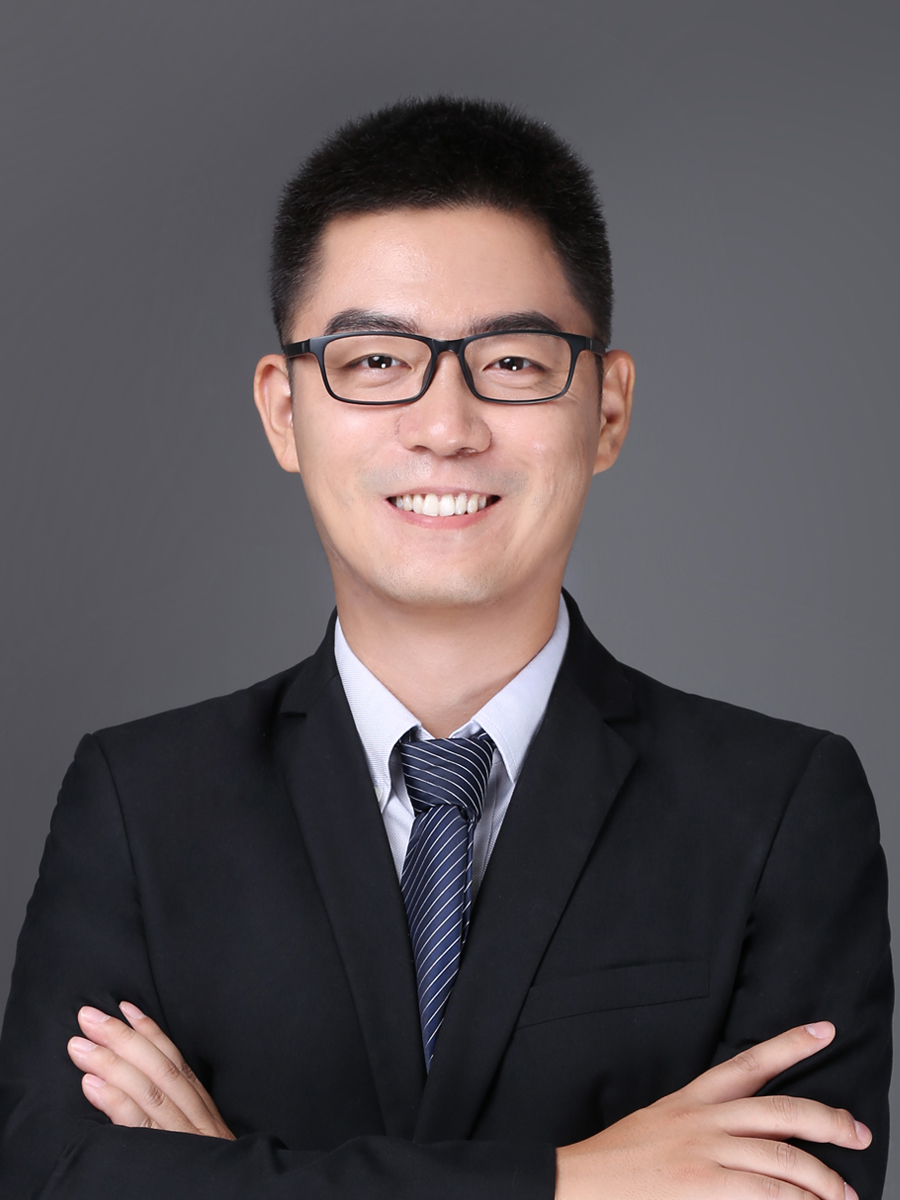}}]
{Boxin Shi}
received the BE degree from the Beijing University of Posts and Telecommunications, the ME degree from Peking University, and the PhD degree from the University of Tokyo, in 2007, 2010, and 2013. He is currently a Boya Young Fellow Assistant Professor and Research Professor at Peking University, where he leads the Camera Intelligence Lab. Before joining PKU, he did research with MIT Media Lab, Singapore University of Technology and Design, Nanyang Technological University, National Institute of Advanced Industrial Science and Technology, from 2013 to 2017. His papers were awarded as Best Paper Runner-Up at ICCP 2015 and selected as Best Papers from ICCV 2015 for IJCV Special Issue. He has served as an associate editor of TPAMI/IJCV and an area chair of CVPR/ICCV. He is a senior member of IEEE.
\end{IEEEbiography}

\end{document}